%% file: main.tex
\documentclass{article} 
\usepackage{iclr2026_conference,times}

\input{math_commands.tex}

\usepackage{hyperref}
\usepackage{url}

\title{%
  Efficient Multi-turn RL for GUI Agents via Decoupled Training and Adaptive Data Curation
}

\iclrfinalcopy

\author{Pengxiang Li$^{1,2}$\thanks{Equal contribution.~~~\Letter~Corresponding author.}~~, Zechen Hu$^{3*}$, Zirui Shang$^{1,2*}$, Jingrong Wu$^{3*}$, Yang Liu$^{2}$, Hui Liu$^{3}$, \\ \textbf{Zhi Gao}$^{1,2\text{\Letter}}$,\textbf{Chenrui Shi}$^{1,2}$, \textbf{Bofei Zhang}$^{2}$, \textbf{Zihao Zhang}$^{3}$, \textbf{Xiaochuan Shi}$^{3}$, \textbf{Zedong Yu}$^{2,4}$, \\ 
\textbf{Yuwei Wu}$^{1,5\text{\Letter}}$, \textbf{Xinxiao Wu}$^{1,5}$, \textbf{Yunde Jia}$^{5}$, \textbf{Liuyu Xiang}$^{4}$, \textbf{Zhaofeng He}$^{4}$, \textbf{Qing Li}$^{2\text{\Letter}}$\\
  \small $^1$Beijing Institute of Technology \small $^2$State Key Laboratory of General Artificial Intelligence, BIGAI \\
\small $^3$DataCanvas  $^4$Beijing University of Posts and Telecommunications  $^5$Shenzhen MSU-BIT University \\ 
\small \url{https://computer-use-agents.github.io/dart-gui}
}

\input{config}

\begin{document}

\maketitle
\vspace{-2em}

\begin{abstract}
Vision-language model (VLM) based GUI agents show promise for automating complex desktop and mobile tasks, but face significant challenges in applying reinforcement learning (RL): (1) slow multi-turn interactions with GUI environments for policy rollout, and (2) insufficient high-quality agent-environment interactions for policy learning.
To address these challenges, we propose \textbf{\acs{dart}}, a \acl{dart} framework for GUI agents, which coordinates heterogeneous modules in a highly decoupled manner. DART separates the training system into four asynchronous modules: environment cluster, rollout service, data manager, and trainer. This design enables non-blocking communication, asynchronous training, rollout-wise trajectory sampling, and per-worker model synchronization, significantly improving the system efficiency: {\textit{1.6$\times$ GPU utilization for rollout}}, {\textit{1.9$\times$ training throughput}}, and {\textit{5.5$\times$ environment utilization}}. 
To facilitate effective learning from abundant samples, we introduce an adaptive data curation scheme: (1) pre-collecting successful trajectories for challenging tasks to supplement sparse success in online sampling; (2) dynamically adjusting rollout numbers and trajectory lengths based on task difficulty; (3) training selectively on high-entropy steps to prioritize critical decisions; (4) stabilizing learning via truncated importance sampling for policy mismatch between policy rollout and updating.
On the OSWorld benchmark, \textit{DART-GUI-7B} achieves a 42.13\% task success rate, a 14.61\% absolute gain over the base model, and 7.34\% higher than open-source SOTA.
\textbf{\textit{We will fully open-source our training framework, data, and model checkpoints}}, which we believe is a timely contribution to the community of agentic RL.
\end{abstract}
\vspace{-3mm}

\begin{figure}[h!]
    \vspace{-2mm}
    \centering
\includegraphics[width=0.95\linewidth]{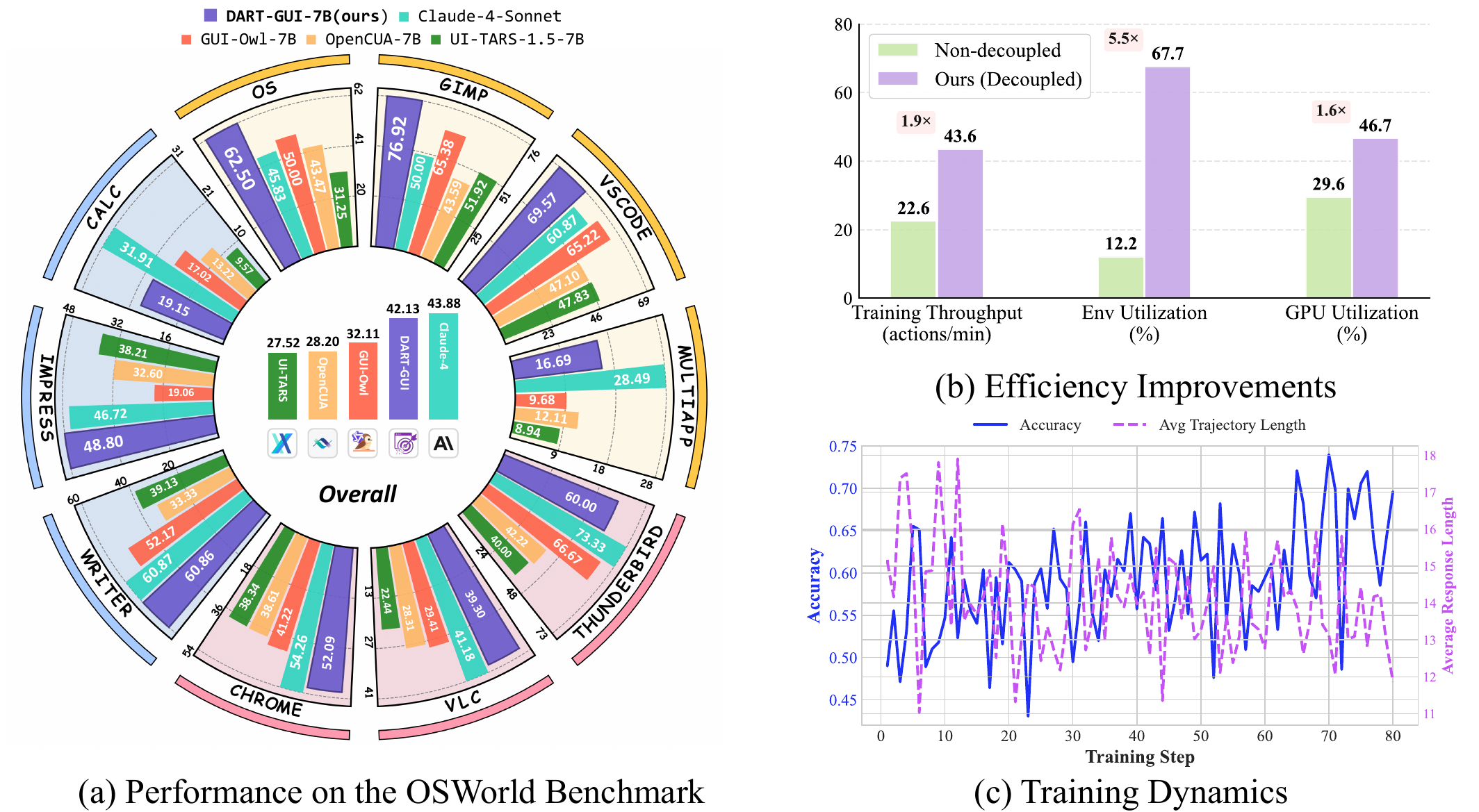}
\vspace{-1mm}
    \caption{{Overview of the \acl{dart} (\textbf{\acs{dart}}) framework for GUI agents.}}
    \label{fig:teaser}
\end{figure}

\section{Introduction}
\vspace{-0.25cm}
The rapid advancement of large language models (LLMs)~\citep{openai2025b,deepseekai2025deepseekr1,gao2024clova} and vision-language models (VLMs)~\cite{bai2025qwen2,wang2024rl,shen2025vlm,gao2024multi,li2025iterative} has accelerated the development of autonomous agents capable of understanding and interacting with graphical user interfaces (GUIs)~\citep{liu2024autoglm,hong2024cogagent}. Such GUI agents~\citep{OpenAI_CUA_2025,bai2025qwen2,guo2025seed1,fu2025mano} hold significant potential for automating complex desktop and mobile tasks by processing screenshots and natural language instructions. While reinforcement learning (RL) has proven effective for enhancing the reasoning and exploration capabilities of LLMs/VLMs in various domains~\citep{wangdistrl,ye2025mobile}, its application to GUI agents remains particularly challenging. GUI tasks typically involve long-horizon, multi-turn interactions that require maintaining context across dozens of states and actions, making RL training processes prohibitively slow and inefficient.

Recent attempts to apply RL for GUI agents~\citep{lu2025arpo,yang2025zerogui,xi2025agentgym} have yielded only modest improvements (2\%–4\% on the OSWorld benchmark~\citep{xie2024osworld}), falling short of closed-source counterparts~\citep{anthropic2025b,OpenAI_CUA_2025,wang2025ui}. We identify two primary bottlenecks: First, the tightly-coupled nature of current RL pipelines, where action prediction, environment interaction, data management, and model updates occur sequentially, creates significant idle time, especially given the long episode lengths typical of GUI tasks. Second, the inherent diversity in task difficulty leads to imbalanced learning: agents may overfit to simpler tasks while struggling to explore successful trajectories for more complex ones. Additionally, the sparse reward signals and the presence of critical decision points within long trajectories can introduce noise and instability during training.

To overcome these limitations, we introduce \textbf{\acs{dart}}, a \acl{dart} framework that decouples the RL process into four specialized, asynchronous modules: environment cluster, rollout service, data manager, and trainer. This design enables non-blocking communication and parallel execution, allowing continuous policy updates alongside ongoing environment interactions. By deploying distributed rollout workers and  rollout-level trajectory sampling, \acs{dart} increases resource utilization, achieving a $1.6\times$ improvement in GPU utilization for rollout, a $1.9\times$ increase in training throughput, and a $5.5\times$ boost in environment utilization compared to coupled baselines.

To further enhance the quality and efficiency of learning, we propose an adaptive data curation strategy that operates at multiple granularities. At the \textit{task} and \textit{trajectory} levels, we pre-collect successful trajectories for challenging tasks and dynamically adjust the number of rollouts and maximum trajectory length based on real-time success rates. At the \textit{step} level, we prioritize training on high-entropy steps, identified as critical decision points, within long trajectories. Finally, at the \textit{token} level, we incorporate a truncated importance sampling term to mitigate distribution shift caused by the inference engine and stabilize policy updates. This curated approach ensures that the agent focuses on the most informative experiences, leading to more robust and efficient learning.

We evaluate our method by training \textit{DART-GUI-7B}, a GUI agent model initialized from UI-TARS-1.5-7B~\citep{qin2025ui}, on the OSWorld benchmark. DART-GUI-7B achieves a 42.13\% task success rate, representing a 14.61\% absolute gain over the base model and a 7.34\% improvement over the previous open-source state-of-the-art. As illustrated in \Cref{fig:teaser}, our framework enables stable performance improvement throughout training, even when exploring with shorter trajectories, and demonstrates superior resource efficiency.

Our main contributions are as follows:
(1) We propose \acs{dart}, a novel decoupled RL framework that significantly enhances training efficiency for GUI agents, achieving $1.6\times$ higher GPU utilization for rollout, $5.5\times$ better environment utilization, and $1.9\times$ higher training throughput.
(2) We introduce an adaptive data curation scheme that optimizes learning at the task, trajectory, step, and token levels, leading to more effective policy updates.
(3) We develop DART-GUI-7B, a state-of-the-art open-source GUI agentic model that achieves superior performance on OSWorld.
To promote reproducibility and advance research in agentic RL, we will \textbf{fully open-source our training framework, model checkpoints, and curated datasets}.

\vspace{-0.25cm}
\section{Related Work}
\vspace{-0.1cm}
\paragraph{GUI Agents}
The development of GUI agents has evolved along three architectural paradigms, each addressing different trade-offs between robustness and generalization. Structured agents~\citep{deng2023mind2web, gur2023real, lai2024autowebglm} leverage metadata such as APIs or accessibility trees to provide semantic clarity and resilience to layout changes, though their effectiveness is inherently bounded by metadata quality and availability. Visual agents~\citep{xu2024aguvis,lu2024omniparser, cheng2024seeclick} directly process raw screenshots through multimodal LLMs, enabling broader applicability across diverse interfaces but introducing sensitivity to visual variations. Hybrid approaches~\citep{wu2024atlas, gou2024navigating, he2024webvoyager} synthesize both modalities, achieving superior grounding performance through complementary information fusion that mitigates the limitations of each individual approach.

The training paradigms for GUI agents have shifted from supervised fine-tuning (SFT)\citep{lin2024showui, qin2025ui, wang2025opencua,zhang2025tongui}, which suffers from limited generalization in complex scenarios, to reinforcement learning approaches that learn from environmental feedback. Early RL methods like GUI-R1\citep{luo2025gui} and InfiGUI-R1~\citep{liu2025infigui} adopted offline training without real-time interaction, struggling with distribution shift and multi-turn reasoning. Recent online RL frameworks address these limitations through various strategies: ARPO~\citep{lu2025arpo} extends GRPO for multi-turn interactions, ZeroGUI~\citep{yang2025zerogui} automates task and reward generation via VLMs, while ComputerRL~\citep{lai2025computerrl} designs asynchronous architectures for training API-equipped GUI Agents. Our work presents a fully open-sourced reinforcement learning framework specifically designed for GUI agents, integrating the decoupled asynchronous framework with adaptive data curation to achieve both superior performance and community accessibility.

\paragraph{Agentic RL}
Reinforcement learning has emerged as a powerful paradigm for enhancing the reasoning and decision-making capabilities of large language models, evolving from preference-based to outcome-based training approaches. While early RLHF methods~\citep{ouyang2022training,rafailov2023direct} relied on expensive human preference annotations that provided only indirect supervision signals, recent advances have shifted toward reinforcement learning with verifiable rewards (RLVR) like GRPO ~\citep{guo2025deepseek} and DAPO~\citep{yu2025dapo}, where automatic and scalable reward signals are derived from concrete task outcomes such as mathematical correctness or code execution success.  GSPO~\citep{xu2024gspo} further improves training stability and efficiency through sequence-level policy optimization that better handles the credit assignment problem in long sequences.

The computational demands of RL have driven the development of asynchronous architectures that decouple different training components for improved efficiency. Building on successful applications in game AI~\citep{vinyals2019grandmaster,berner2019dota}, recent frameworks have adapted asynchronous training for language models: AREAL~\citep{fu2025areal} separates rollout generation from model training to maximize GPU utilization while employing staleness-aware PPO to maintain training stability; ROLL~\citep{wang2025reinforcement} provides a comprehensive RL library supporting multi-model pipelines and flexible resource scheduling for large-scale LLM training. These frameworks focus on general text-based or multimodal tasks with relatively dense rewards and shorter interaction sequences. We present a decoupled RL framework specifically designed for GUI agent training, addressing the unique challenges of long-horizon multi-modal interactions.

\section{\acs{dart}: \acl{dart} Framework}
\label{sec:dart}

\subsection{Formulation}

We formulate GUI tasks as a sequential decision-making process. At the time step $t$, given the current visual state $s_t$ (a screenshot of GUI) and the interaction history $h_t = \{(s_{\mathrm{max}(1,t-m}), t_{\mathrm{max}(1,t-m)}, a_{\mathrm{max}(1,t-m)}), \ldots, (s_{t-1}, r_{t-1}, a_{t-1})\}$ of previous $m$ steps, where $r$ denotes the thought for reasoning and $a$ denotes the action (such as clicking on a specific UI element or entering text), along with the task $\tau$, the agent generates a new thought $r_t$ and an executable action $a_t$. Executing the action $a_t$ leads to a new visual state $s_{t+1}$ (an updated screenshot). This interaction loop continues, with the agent repeatedly observing the environment, producing thought and actions, and receiving updated observations until either a termination condition is met (\textit{e.g.}, the task is completed or fails) or the maximum number of steps is reached. We parameterize the GUI agent using a policy model $\pi_\theta$ (\textit{i.e.}, an VLM) that generates thoughts and actions based on the current state and historical context: $r_t^*, a_t^* = \arg\max_{r_t, a_t} \pi_\theta(a_t | \tau, h_t, s_t)$, where $r_t^*$ and $a_t^*$ represents the optimal thought and action selected by the policy model.

\begin{figure}
    \vspace{-1cm}
    \centering
    \includegraphics[width=0.9\linewidth]{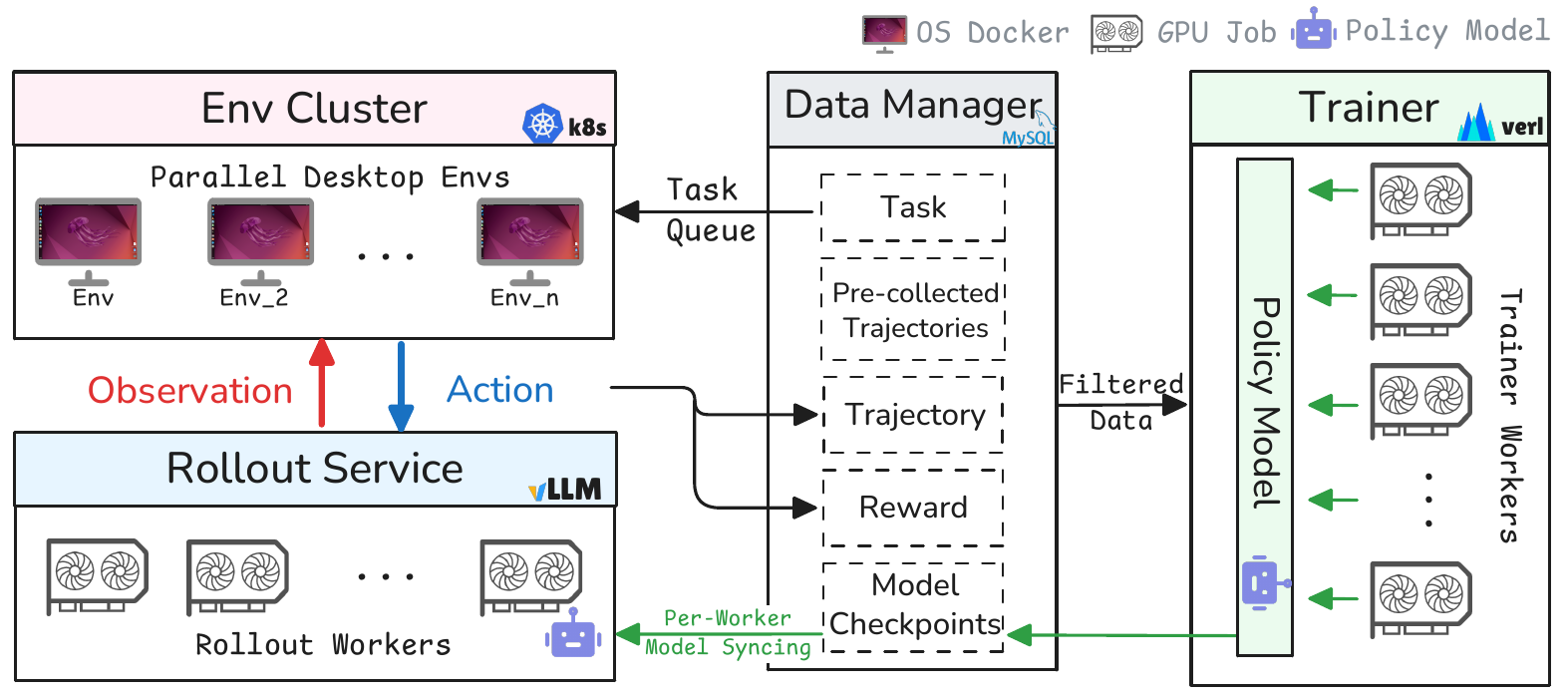}
    \vspace{-0.5em}
    \caption{Overall architecture of our framework. The \textit{Rollout Service} interacts with multiple environments in parallel to generate trajectories, which are managed and delivered to the \textit{Trainer} for policy updates. Updated actors are synchronized back to the \textit{Rollout Service}, enabling scalable and efficient asynchronous learning. Implementation techniques are annotated within the figure.}
    \label{fig:arch}
\end{figure}

\vspace{-0.5em}
\subsection{Architecture}
\vspace{-0.5em}

Our RL framework contains four decoupled modules: \textit{Trainer}, \textit{Data Manager}, \textit{Env Cluster}, and \textit{Rollout Service}, where none of them will be blocked by other modules, as shown in~\Cref{fig:arch}.
We set up hundreds of real desktop environments in \textit{Env Cluster}, and design a \textit{Rollout Service} to load multiple policy models.
The \textit{Env Cluster} receives a sequence of tasks from the \textit{Data Manager}, and samples $N$ trajectories for each task.
The \textit{Rollout Service} dynamically assigns idle workers to produce thoughts and actions for different rollouts in parallel.
We store sampled trajectories and corresponding rewards in the \textit{Data Manager}.
When $N$ trajectories of one task are finished, the \textit{Data Manager} filters and passes them to the \textit{Trainer} based on predefined rules for policy updates.
Finally, the updated policy model is synchronized back to the \textit{Rollout Service}, enabling scalable and highly efficient asynchronous learning. 
Key interactions among these modules are as follows.

\vspace{-0.5em}
\subsection{Asynchronous Trainer}
\vspace{-0.5em}

To improve GPU and environment utilization, we decouple the \textit{Trainer} from the trajectory rollout process, which avoids the blocking between training and rollout.
The \textit{Trainer} operates asynchronously, receiving filtered trajectories from the \textit{Data Manager} and performing step-wise GRPO updates. The updated model weights are synchronized to the \textit{Rollout Service}, enabling continuous training while new trajectories are sampled simultaneously.

\textbf{Step-wise GRPO.} 
We adopt step-wise Group Relative Policy Optimization (GRPO) \citep{shao2024deepseekmath} to train our GUI agent. For each task $\tau$, we sample $N$ trajectories $T_1, T_2, \ldots, T_N$ using the current policy $\pi_{\theta_{\text{old}}}^{\mathrm{Rollout}}$ of the \textit{Rollout Service}, where the $i$-th trajectory has length $L_i$ and consists of state-thought-action pairs: $T_i = \{(s_{i,j}, r_{i,j}, a_{i,j})\}_{j=1}^{L_i}$. Each trajectory receives a reward $R_i$, and we decompose each trajectory into individual steps and group all steps from the same task for advantage computation. Specifically, we create a step group $\mathcal{D} = \{(h_{i,j}, s_{i,j}, r_{i,j}, a_{i,j}, R_i) \mid i \in [1,N], j \in [1,L_i]\}$, where each step $(s_{i,j}, r_{i,j}, a_{i,j})$ is combined with its history $h_{i,j}$ and the trajectory-level reward $R_i$. We denote ``$h_{i,j}, s_{i,j}, r_{i,j}, a_{i,j}$'' as ``$h,s,r,a$'' for simplicity. 
The step-wise GRPO objective is formulated as

\begin{equation}
\scalebox{0.7}{$
\begin{aligned}
\mathcal{J}(\theta) &= \mathbb{E}_{(h,s,a,R) \sim \mathcal{D}} \left[  \nabla_\theta \min\left( \frac{\pi^{\mathrm{Train}}_\theta(a|h,s)}{\pi^{\mathrm{Train}}_{\text{old}}(a|h,s)}A,  \text{clip}\left(\frac{\pi^{\mathrm{Train}}_\theta(a|h,s)}{\pi^{\mathrm{Train}}_{\text{old}}(a|h,s)}, 1-\epsilon_{\text{low}}, 1+\epsilon_{\text{high}}\right)A\right) -\,\beta\, D_{\mathrm{KL}}\!\big(\pi_\theta^{\mathrm{Train}}(a|h,s)\,\|\,\pi_{\theta}^{\text{Ref}}(a| h,s)\big)
\right],
\end{aligned}
$}
\label{eq:lossfunction}
\end{equation}

where the advantage is computed as
\[
\scalebox{0.87}{$
A_{i,j} = \frac{R_i - \bar{R}}{\sigma_R}, \quad 
\bar{R} = \frac{1}{|\mathcal{D}|}\sum_{(h,s,a,R) \in \mathcal{D}} R, \quad 
\sigma_R^2 = \frac{1}{|\mathcal{D}|}\sum_{(h,s,a,R) \in \mathcal{D}} (R - \bar{R})^2.
$}
\]

\subsection{Rollout-wise Sampling}

\begin{figure}[t!]
    \vspace{-1cm}
    \centering
    \includegraphics[width=1.0\linewidth]{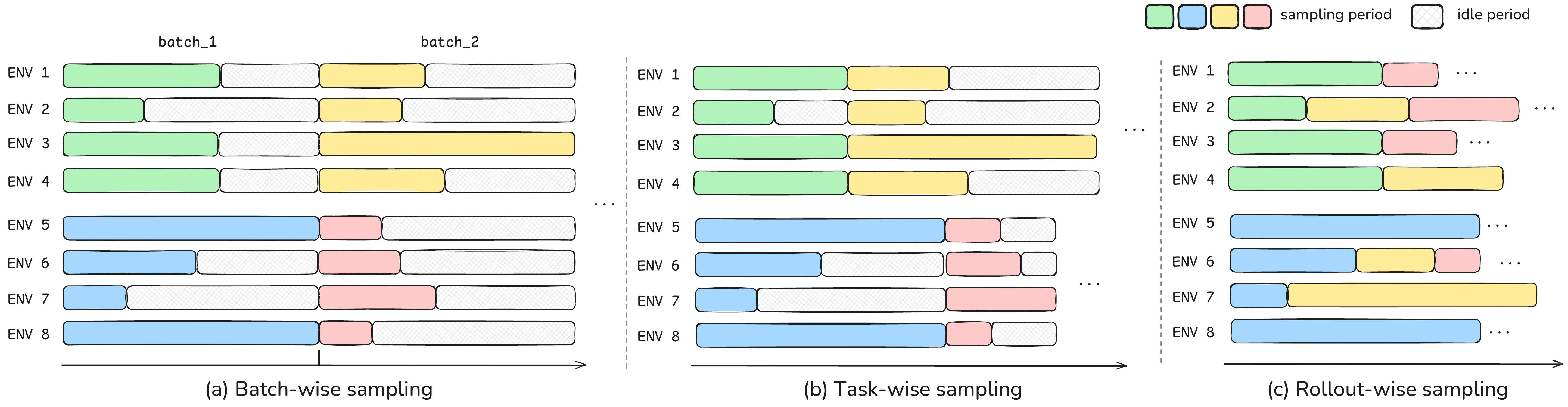}
    \vspace{-0.5em}
    \caption{Visualization of sampling timelines for 4 tasks in two batches (denoted by different colors) with rollout number $N=4$, batch size $=2$, and a total of $8$ environments. Each bar represents the timeline of rollout execution of a task on one environment. }
    \label{fig:sampling}
\end{figure}

Solving practical GUI tasks (such as OSWorld) 
typically consists of dozens of steps and lasts for tens of minutes.
Thus, sampling efficiency often becomes a major bottleneck. 
When sampling, tasks within the same batch may vary significantly in difficulty, and even for the same task, minor environmental variations across different executions may produce trajectories of vastly different lengths\citep{wang2025opencuaopenfoundationscomputeruse}. 
Under conventional batch-wise sampling (\Cref{fig:sampling} (a)), this heterogeneity causes substantial resource underutilization: environments that complete early remain idle while GPUs stay underused until all trajectories in the batch finish. Task-wise sampling (\Cref{fig:sampling} (b)) partially addresses this issue but still requires waiting for entire tasks to complete before resuming new sampling, limiting overall efficiency. We implement rollout-wise sampling (~\Cref{fig:sampling} (c)), where an individual trajectory serves as the minimal scheduling unit. Once an environment completes a rollout, it immediately launches the next sampling request without waiting for others. This fine-grained scheduling significantly improves environment utilization and maximizes GPU throughput.

Furthermore, rather than statically assigning environments to fixed rollout workers, our decoupled training framework introduces a dynamic model service pool for load balancing. All GPUs are pooled into the shared \textit{Rollout Service}, with incoming requests being distributed to works based on current device utilization. This design provides two engineering advantages: (1) all environments communicate to the \textit{Rollout Rervice} through a unified interface, simplifying system architecture and coordination; and (2) balanced GPU workloads minimize idle time and bottlenecks, resulting in faster inference and higher overall throughput.

\subsection{Per-Worker Model Synchronization}

\begin{figure}[t]
    \vspace{-1cm}
    \centering
    \includegraphics[width=1.0\linewidth]{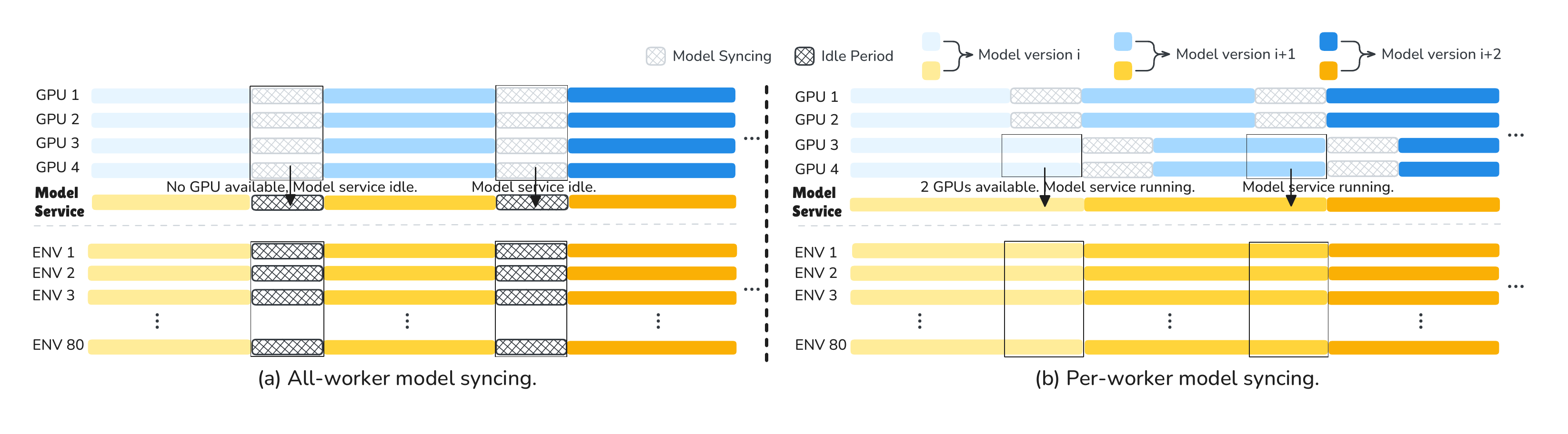}
    \vspace{-2em}
    \caption{Timeline comparison between all-worker and per-worker model updating for 4 GPUs and 80 environments. The timelines depict idle periods for the GPUs and environments across two model updates. Different model versions are represented by varying shades of color. For per-worker model updating, the device number per worker is set to 2.}
    \label{fig:update}
\end{figure}

Model synchronization presents a critical bottleneck in asynchronous GUI agent RL. Traditional approaches rely on global synchronization: when the \textit{Trainer} completes a training iteration, all rollout workers halt operations and wait for every GPU device to receive the updated weights before resuming sampling. As shown in \Cref{fig:update} (a), this creates system-wide downtime where environments sit idle and GPUs remain underutilized during each update. 

We introduce per-worker model update to eliminate this bottleneck through staggered parameter distribution. Instead of synchronized model updates across all rollout workers simultaneously, we gradually refresh model weights on rollout workers. It means that when one worker is updating model weights, the others continue serving inference requests with its current model version. \Cref{fig:update} (b) illustrates how this maintains continuous service availability: environments never experience complete blocking. This approach delivers two key advantages: dramatically improved sampling throughput through reduced idle time, and seamless model version transitions that preserve ongoing rollout stability.

\section{Multi-Level Adaptive Data Curation for GUI Tasks}
\label{sec:data}

\subsection{Performance-Aware Task Rollout}

Fixed sampling strategies in reinforcement learning waste computational resources by treating all tasks equally. Easy tasks receive excessive sampling while difficult tasks lack sufficient exploration. We propose performance-aware task rollout that dynamically adjusts both sampling frequency and trajectory length based on each task's learning progress.

\begin{wrapfigure}{r}{0.26\linewidth}
    \centering
    \vspace{-12pt}
    \includegraphics[width=\linewidth]{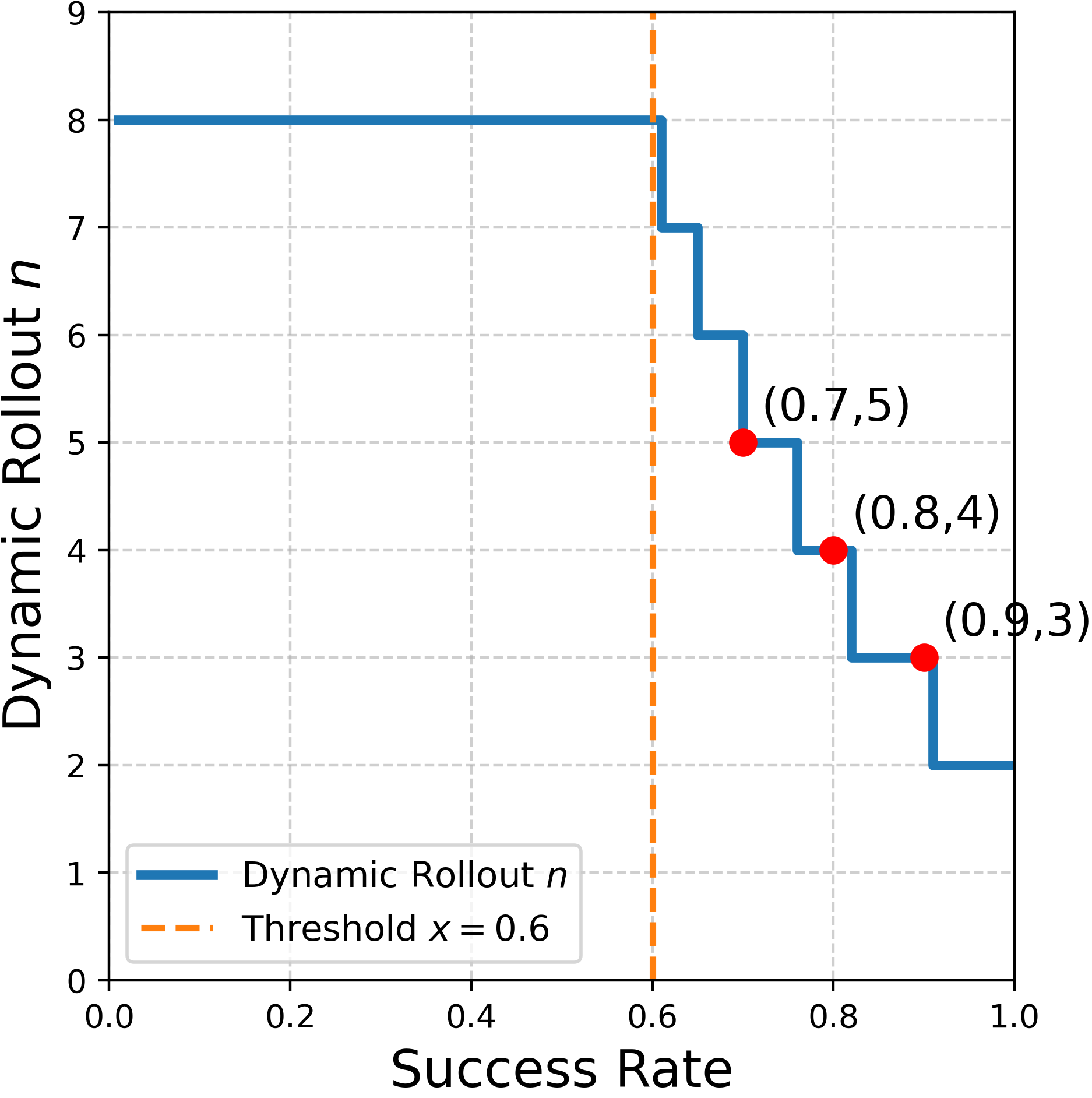}
    \caption{Dynamic rollout $N$ with task success rate.}
    \label{fig:dynamic_rollout}
    \vspace{-18pt}
\end{wrapfigure}

\textbf{Dynamic Rollout Frequency}
We continuously monitor each task's success rate and adjust its sampling frequency accordingly. As shown in Figure~\ref{fig:dynamic_rollout}, when a task achieves high success rates (above 0.6), we reduce its rollout frequency from 8 to lower values, preventing overfitting on already-solved tasks. Tasks with low success rates maintain maximum sampling to ensure adequate learning opportunities. This strategy reallocates computational resources from well-learned tasks to challenging ones, improving overall training efficiency.

\textbf{Dynamic Trajectory Length}
We set task-specific trajectory length limits based on the historical maximum length of successful completions. Instead of using a fixed limit for all tasks, each task receives its own length threshold derived from its successful trajectories. This prevents wasting computation on hopeless long trajectories while allowing sufficient exploration for tasks that genuinely require more steps. For instance, simple clicking tasks might terminate after 10 steps, while complex multi-application tasks can extend to 50 steps.
This adaptive approach optimizes the balance between thorough exploration and computational efficiency for individual tasks.

\subsection{Experience Pool of Trajectories}

Training on challenging tasks poses a significant obstacle in reinforcement learning due to extremely low success rates, which results in insufficient positive trajectories in rollouts. 
It is common that
all trajectories of a task fail, providing no effective learning signal for policy improvement. This severe imbalance between simple tasks and difficult tasks results in training instability, preventing the model from learning correct behavioral patterns. To address this limitation, we introduce an \textit{Experience Pool} that serves as a repository of high-quality successful trajectories for challenging tasks, enabling dynamic supplementation for difficult tasks during training to ensure balanced learning signals.

Specifically, we pre-populate the \textit{Experience Pool} by collecting and storing high-quality successful trajectories through preliminary sampling. During training, when the system detects that all trajectories in the current task fail, it automatically triggers the pool sampling mechanism to randomly retrieve a successful trajectory and incorporate it into the training batch. This design guarantees that every training task contains at least one positive sample while maintaining a reasonable balance between exploration and experience replay, thereby preventing performance degradation on challenging tasks and improving overall training stability.

\subsection{High-Entropy-Driven Step Optimization}

Inspired by \citet{wang2025beyond}, who showed that training exclusively on high-entropy tokens that "act as critical forks that steer the model towards diverse reasoning pathways" can drive effective reinforcement learning, we extend this insight to multi-turn GUI agent training.
Low entropy means non-critical steps for a GUI task, and using such steps may cause instability in training.
We design an entropy-based step selection mechanism that identifies and prioritizes the top 80\% high-entropy steps for training, thereby encouraging exploration in critical decision-making moments.

For the $t$-th step in a trajectory, we calculate the step-level entropy $H_t$ as the average entropy across all tokens generated in the concatenated thought $r_t$ and action $a_t$ sequence: $H_t = \frac{1}{|r_t| + |a_t|} \sum_{i=1}^{|r_t| + |a_t|} H_{t,i}$,
where $H_{t,i} = -\sum_{v=1}^{V} p_{t,i,v} \log p_{t,i,v}$ represents the token-level entropy, with $p_{t,i,v} = \pi_\theta(v | \tau, h_t, s_t, o_{<i})$ being the probability of generating the token $v$.
During training, we modify the GRPO objective to include only steps whose entropy is at least larger than $20\%$ steps within the group.
This makes reinforcement learning focuses on uncertain steps in GUI navigation.


\subsection{Distribution Alignment for OOD Tokens}

During training, the discrepancy between quantization strategies employed by the \textit{Rollout Service} and the \textit{Trainer} leads to significant differences in their generated policy distributions. In addition, the pre-collected trajectories also has different distributions from the current model.
To solve this issue, we follow \citet{yao2025offpolicy} and incorporate a truncated importance sampling weight $\min\left(\frac{\pi^\mathrm{Train}_{\theta_\text{old}}(a|h,s)}{\pi^{\mathrm{Rollout}}_{\theta_{\text{old}}}(a|h,s)}, C\right)$ into the training objective in Eq.(\ref{eq:lossfunction}) to mitigate this gap. 
By reweighting the gradient contributions based on the probability ratio between the two distributions, we utilize unbiased learning for our decoupled framework, enabling stable training. The final objective is
\begin{equation}
\scalebox{0.85}{$
\begin{aligned}
\mathcal{J}_{\text{HE}}(\theta) &= \mathbb{E}_{(h,s,a,R) \sim \mathcal{D}} \Bigg[ 
\mathbb{I}[H_t \geq \tau_{\mathcal{D}}^{0.2}] \cdot \bigg( 
\min\left(\frac{\pi^\mathrm{Train}_{\text{old}}(a|h,s)}{\pi^{\mathrm{Rollout}}_{\text{old}}(a|h,s)}, C\right) \cdot
\nabla_\theta \min\Bigg( \frac{\pi^{\mathrm{Train}}_\theta(a|h,s)}{\pi^{\mathrm{Train}}_{\text{old}}(a|h,s)}A,  \\
& \text{clip}\!\left(\frac{\pi^{\mathrm{Train}}_\theta(a|h,s)}{\pi^{\mathrm{Train}}_{\text{old}}(a|h,s)}, 1-\epsilon_{\text{low}}, 1+\epsilon_{\text{high}}\right)A\Bigg) -\,\beta\, D_{\mathrm{KL}}\!\big(\pi_\theta^{\mathrm{Train}}(a|h,s)\,\|\,\pi_{\theta}^{\text{Ref}}(a| h,s)\big)\bigg)
\Bigg]
\end{aligned}
$}
\end{equation}
where $\mathbb{I}[\cdot]$ is the indicator function, and $\tau_{\mathcal{D}}^{0.2}$ is the threshold larger than $20\%$ entropy in the group.



\section{Experiment}

\subsection{Settings} \label{subsec:imple}
\label{subsec:setting}

\paragraph{Experimental Setup} 
We evaluate our approach on OSWorld-Verified \citep{xie2024osworld}, a comprehensive benchmark for assessing multimodal autonomous agents in realistic computer environments. 
For our training corpus, we adopt the sampling methodology proposed by \citet{lu2025arpo}, selecting a representative subset of 203 tasks from the OSWorld benchmark. We use the results of the evaluation scripts from the OSWorld as rewards for the RL.

\paragraph{Evaluation Protocol}
We follow the OSWorld evaluation framework \citep{xie2024osworld}, which uses execution-based validation scripts to assess task completion. Each trajectory receives a reward score in [0, 1] based on programmatic verification of the final system state against predefined success criteria.

\paragraph{Implementation}
We adopt UI-TARS-1.5-7B~\citep{qin2025ui} as the baseline for our policy model.
Unlike approaches that rely on multi-agent systems, agent workflows, or agents equipped with additional APIs/tools, we focus on enhancing the capabilities of a single VLM agent through reinforcement learning, aiming to improve the model's inherent decision-making abilities without external scaffolding. 
Based on decoupled agentic RL training (DART) and the data curation scheme, we obtain the DART-GUI-7B model. 
More details about training and RL framework can be found in Appendix~\ref{app:implementation}.

\subsection{Main Results}
\label{subsec:result}

Table~\ref{tab:comp_use_res} presents results on the OSWorld benchmark across 10 diverse applications. 
Notably, \ac{dart}-GUI-7B demonstrates superior sample efficiency compared to both open-source and closed-source models.
DART-GUI-7B achieves 42.13\% overall success rate with only 30 maximum steps, establishing a new state-of-the-art among open-source models and showing a 12.71\% improvement over the baseline UI-TARS-1.5-7B (27.52\% with 100 steps). 
It also achieves comparable performance to Claude-4-Sonnet (41.39\% with 100 steps) and outperforming models like OpenAI CUA o3 (23.00\% with 100 steps).
This significant gain validates the effectiveness of our decoupled asynchronous RL framework and multi-level data curation strategies that focuses on critical decision points. \ac{dart}-GUI-7B shows consistent improvements across all applications, with particularly strong gains in complex system-level tasks: OS tasks improve by 31.25\% (62.50\% vs. 31.25\%), LibreOffice Writer by 21.73\% (60.86\% vs. 39.13\%), and Thunderbird by 20.00\% (60.00\% vs. 40.00\%). These applications involve longer interaction sequences and diverse action spaces, highlighting our framework's ability to handle long-horizon tasks with sparse rewards effectively.


\setlength{\tabcolsep}{2pt}
\begin{table}[t]
    \vspace{-1cm}
    \centering
    \scriptsize
    \caption{Results on the OSWorld benchmark. Max Steps indicates the maximum number of agent-environment interactions allowed. \textbf{Bold} values denote the best performance among \textit{open-source} models. For brevity, LibreOffice Calc, Impress, and Writer are abbreviated as \textit{calc}, \textit{impress}, and \textit{writer}, respectively. Our results are obtained through evaluation on self-deployed devices using the official codebase and Docker environment. * means self-reported results in the method.
    }
    \vspace{-1em}
    \resizebox{1\textwidth}{!}{%
    \begin{tabular}{l|c|cccccccccc|c}
    \hline
     \multirow{2}{*}{\textbf{Model}} & \multirow{2}{*}{\textbf{Max Steps}} & \multicolumn{11}{c}{\textbf{Task Success Rate (\%)}} \\
     &  & \textit{chrome} & \textit{gimp} & \textit{calc} & \textit{impress} & \textit{writer} & \textit{multi\_apps} & \textit{os} & \textit{thunderbird} & \textit{vlc} & \textit{vs\_code} & \textit{Overall} \\
    \hline
    \rowcolor{customgray} \multicolumn{13}{c}{\textit{Closed-source Model}} \\ \hline
    OpenAI CUA o3~\citep{openai2025b} & 50 & 21.74 & 38.46 & 8.51 & 4.26 & 21.74 & 11.83 & 37.50 & 20.00 & 17.59 & 21.74 & 17.17 \\
    OpenAI CUA o3~\citep{openai2025b} & 100 & 13.04 & 38.46 & 10.64 & 10.64 & 30.43 & 16.53 & 62.50 & 26.67 & 39.18 & 39.13 & 23.00 \\
    TianXi-Action-7B~\citep{tang2025sea} & 50 & 36.83 & 55.77 & 6.38 & 38.24 & 54.35 & 6.60 & 38.22 & 43.33 & 31.85 & 67.39 & 29.81 \\
    OpenAI CUA~\citep{OpenAI_CUA_2025} & 100 & 33.61 & 48.08 & 18.09 & 24.47 & 23.91 & 15.61 & 54.17 & 56.67 & 38.21 & 60.87 & 30.50 \\
    OpenAI CUA~\citep{OpenAI_CUA_2025} & 50 & 36.87 & 34.62 & 14.89 & 29.70 & 26.09 & 15.81 & 70.83 & 66.67 & 11.76 & 69.57 & 31.30 \\
    Claude-3-7-Sonnet~\citep{anthropic2025a} & 100 & 54.26 & 42.31 & 21.28 & 31.83 & 47.83 & 19.62 & 45.83 & 66.67 & 24.88 & 56.52 & 35.57 \\ 
    Claude-3-7-Sonnet~\citep{anthropic2025a} & 50 & 52.09 & 38.46 & 31.91 & 36.09 & 43.48 & 17.66 & 50.00 & 53.33 & 23.53 & 56.52 & 35.83 \\
    DeepMiner-Mano-7B~\citep{fu2025mano} & 100 & 39.13 & 69.23 & 27.66 & 42.47 & 56.52 & 17.20 & 50.00 & 73.33 & 35.29 & 78.26 & 40.16 \\
    Seed1.5-VL-250717~\citep{guo2025seed1} & 100 & 56.52 & 50.00 & 34.78 & 48.91 & 56.52 & 15.35 & 39.13 & 73.33 & 35.29 & 56.52 & 40.18 \\
    Claude-4-Sonnet~\citep{anthropic2025b} & 100 & 47.74 & 50.00 & 29.79 & 42.47 & 52.17 & 27.86 & 45.83 & 66.67 & 32.82 & 69.57 & 41.39 \\
    UI-TARS-250705~\citep{wang2025ui} & 100 & 56.43 & 50.00 & 40.43 & 55.30 & 60.87 & 14.66 & 41.67 & 66.67 & 44.00 & 52.17 & 41.84 \\
    Claude-4-Sonnet~\citep{anthropic2025b} & 50 & 54.26 & 50.00 & 31.91 & 46.72 & 60.87 & 28.49 & 45.83 & 73.33 & 41.18 & 60.87 & 43.88 \\    
    \hline
    \rowcolor{customgray} \multicolumn{13}{c}{\textit{Open-Source Model}} \\ \hline
    Qwen2.5-VL-32B~\citep{bai2025qwen2} & 100 & 8.70 & 3.85 & 0.00 & 0.00 & 8.70 & 2.15 & 8.33 & 6.67 & 0.00 & 8.70 & 3.88 \\
    Qwen2.5-VL-72B~\citep{bai2025qwen2} & 100 & 4.35 & 0.00 & 6.38 & 0.00 & 8.70 & 3.23 & 16.67 & 13.33 & 5.88 & 4.35 & 4.99 \\
    ZeroGUI*~\citep{yang2025zerogui} & 15 & - & - & - & - & - & - & - & - & - & - & 20.2 \\
    UI-TARS-72B-dpo~\citep{qin2025ui} & 50 & 33.24 & 61.54 & 12.77 & 25.45 & 43.48 & 6.71 & 33.33 & 33.33 & 23.53 & 47.83 & 25.88 \\
    OpenCUA-7B~\citep{wang2025opencua} & 100 & 36.14 & 47.44 & 10.64 & 31.90 & 27.54 & 9.87 & 42.87 & 40.00 & 29.41 & 44.93 & 26.60 \\
    UI-TARS-72B-dpo~\citep{qin2025ui} & 100 & 32.61 & 73.08 & 6.38 & 23.81 & 34.78 & 8.29 & 37.50 & 60.00 & 17.65 & 52.17 & 26.84 \\
    UI-TARS-1.5-7B~\citep{qin2025ui} & 50 & 32.79 & 53.85 & 8.51 & 39.31 & 41.30 & 8.60 & 25.54 & 46.67 & 24.41 & 52.17 & 27.25 \\
    UI-TARS-1.5-7B~\citep{qin2025ui} & 100 & 38.34 & 51.92 & 9.57 & 38.21 & 39.13 & 8.94 & 31.25 & 40.00 & 22.44 & 47.83 & 27.52 \\
    OpenCUA-7B~\citep{wang2025opencua} & 50 & 38.61 & 43.59 & 13.22 & 32.60 & 33.33 & 12.11 & 43.47 & 42.22 & 28.31 & 47.10 & 28.20 \\
    ARPO*~\citep{lu2025arpo} & 15 & - & - & - & - & - & - & - & - & - & - & 29.9\\
    GUI-Owl-7B~\citep{ye2025mobile} & 15 & 41.22 & 65.38 & 17.02 & 19.06 & {52.17} & 9.68 & 50.00 & \textbf{66.67} & 29.41 & 65.22 & 32.11 \\
    OpenCUA-32B~\citep{wang2025opencua} & 50 & 43.39 & {69.23} & 13.48 & 38.28 & 40.58 & 14.93 & 53.62 & 53.33 & 25.49 & 55.07 & 34.14 \\
    OpenCUA-32B~\citep{wang2025opencua} & 100 & 39.77 & 66.67 & 18.44 & 37.60 & 36.23 & {16.21} & 55.07 & 46.67 & 33.33 & 63.31 & 34.79 \\
\hline
\rowcolor{customgreen} \multicolumn{13}{c}{\textit{Baseline}} \\
\hline
UI-TARS-1.5-7B & 100 & 38.34 & 51.92 & 9.57 & 38.21 & 39.13 & 8.94 & 31.25 & 40.00 & 22.44 & 47.83 & 27.52 \\

\hline
\rowcolor{customblue} \multicolumn{13}{c}{\textit{Ours}} \\
\hline

DART-GUI-7B & 30 & \textbf{52.09} & \textbf{76.92} & \textbf{19.15} & \textbf{48.80} & \textbf{60.86} & \textbf{16.69} & \textbf{62.50} & 60.00 & \textbf{39.30} & \textbf{69.57} & \textbf{42.13} \\

\rowcolor{gray!8}
$\Delta$ (Ours \emph{vs.} Baseline)&- & ${\color{red}\uparrow}{\tiny\textcolor{gray}{13.75}}$ & ${\color{red}\uparrow}{\tiny\textcolor{gray}{25.00}}$ & ${\color{red}\uparrow}{\tiny\textcolor{gray}{9.58}}$ & ${\color{red}\uparrow}{\tiny\textcolor{gray}{10.59}}$ & ${\color{red}\uparrow}{\tiny\textcolor{gray}{21.73}}$ & ${\color{red}\uparrow}{\tiny\textcolor{gray}{7.75}}$ & ${\color{red}\uparrow}{\tiny\textcolor{gray}{31.25}}$ & ${\color{red}\uparrow}{\tiny\textcolor{gray}{20.00}}$ & ${\color{red}\uparrow}{\tiny\textcolor{gray}{16.86}}$ & ${\color{red}\uparrow}{\tiny\textcolor{gray}{21.74}}$ & ${\color{red}\uparrow}{\tiny\textcolor{gray}{14.61}}$ \\

    \hline
    \end{tabular}%
    }
    \label{tab:comp_use_res}
\end{table}

\subsection{Efficiency Analysis}

We evaluate our decoupled framework's efficiency gains across three key metrics, as shown in Table~\ref{tab:efficiency}. Our framework achieves substantial improvements: training throughput nearly doubles from 22.6 to 43.6 actions/min (1.9$\times$), environment utilization increases dramatically from 12.2\% to 67.7\% (5.5$\times$), and GPU utilization improves from 29.6\% to 46.7\%(1.6$\times$). These gains stem from our decoupled design's elimination of system-wide blocking. Environments continuously generate rollouts without waiting for batch completion, enabling immediate trajectory generation upon task completion. Worker-wise model updates avoid global synchronization, allowing the GPU service to continuously perform inference while some are updating models. This asynchronous operation minimizes idle time across all components, demonstrating that decoupling is essential for efficient GUI agent RL at scale.

\setlength{\tabcolsep}{4pt}
\begin{table}[htbp]
    \centering
    \scriptsize
    \caption{Efficiency improvements of our decoupled framework compared to non-decoupled baseline.}
    \resizebox{0.75\textwidth}{!}{%
    \begin{tabular}{lccc}
        \hline
        \textbf{System Setup} & \textbf{Training Throughput (actions/min)} & \textbf{Env Util. (\%)} & \textbf{GPU Util. (\%)} \\
        \hline
        Non-Decoupled & 22.6 & 12.2 & 29.6 \\
        Decoupled (Ours) & 43.6 & 67.7 & 46.7 \\
        \hline
        \rowcolor{gray!5} \textbf{Improvement} & \textbf{1.9$\times$} & \textbf{5.5$\times$} & \textbf{1.6$\times$} \\
        \hline
    \end{tabular}%
    }
    \label{tab:efficiency}
    \vspace{-10pt}
\end{table}

\begin{figure}[t!]
    \vspace{-1cm}
    \centering
    \includegraphics[width=1.0\linewidth]{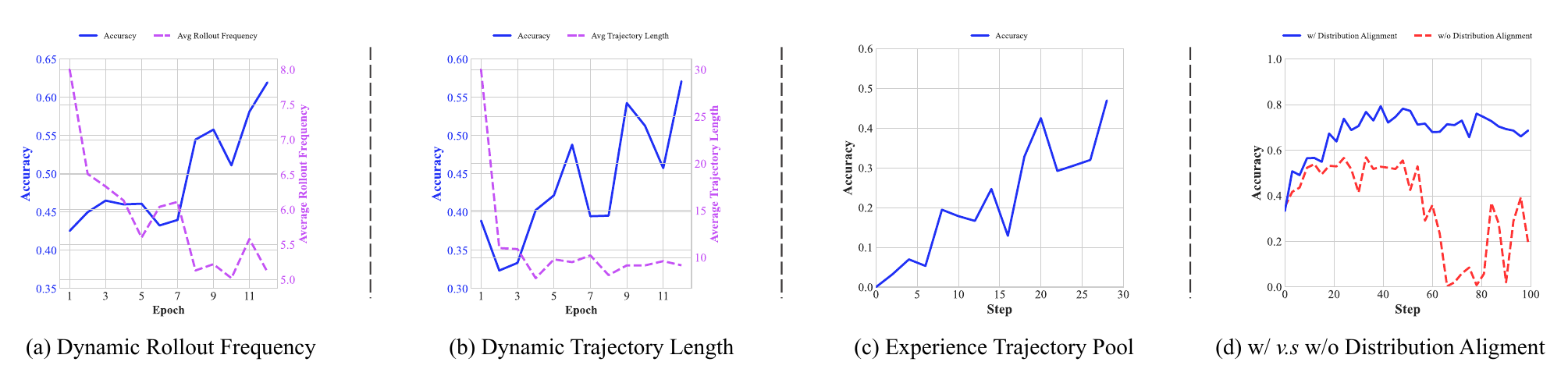}
    \vspace{-1.5em}
    \caption{(a) Dynamic rollout frequency vs. model accuracy across epochs. (b) Dynamic trajectory length vs. model accuracy over training. (c) Impact of experience trajectory pool on accuracy. (d) Performance comparison with and without distribution alignment.}
    \vspace{-1.0em}
    \label{fig:ablation}
\end{figure}

\begin{table}[htbp]
    \centering
    \caption{Ablation study of the data curation scheme. 
    DR stands for dynamic rollout, DTL for dynamic trajectory length, HE for high-entropy-driven step selection, and DA for distribution alignment.}
    \label{tab:ablation}
    \resizebox{0.6\linewidth}{!}{%
    \begin{tabular}{lccccc|c}
        \toprule
        & \textbf{Baseline} & \textbf{w/ DR} & \textbf{w/ DTL} & \textbf{w/ HE} & \textbf{w/ DA} & \textbf{Ours} \\
        \midrule
        \textbf{Pass@1 (\%)} & 28.67 & 50.90 & 66.11 & 68.33 & 70.55 & \textbf{72.28} \\
        \bottomrule
    \end{tabular}%
    }
\end{table}

\vspace{-1cm}
\subsection{Ablation}
\label{subsec:ablation}
We conduct ablation studies on a subset of 45 tasks from the training set to evaluate the four levels of our data curation scheme. Baseline means only using the decoupled RL training framework, and Ours means we apply the whole data curation scheme is added.



\textbf{Dynamic Rollout and Dynamic Trajectory Length.}
We evaluate the impact of our mechanisms on training efficiency. As shown in \Cref{tab:ablation}, both approaches effectively improves the baseline performance. As shown in ~\Cref{fig:ablation}(a) and (b), both strategies effectively reduce computational overhead as the model improves. Dynamic Rollout demonstrates that as accuracy increases, the average rollout frequency decreases from 8.0 to 5.0 per task.
Similarly, Dynamic Trajectory Length shows that as performance improves from, the average trajectory length drops from 30 to less than 10 steps, as the model learns to complete tasks more efficiently.
This confirms that our adaptive mechanisms successfully accelerate training by eliminating redundant computation while maintaining exploration on challenging tasks.

\textbf{Experience Pool of Trajectories.}
We evaluate the impact of the experience pool on training performance for a set of 22 challenging tasks which initially exhibits a success rate of 0\%. As illustrated in ~\Cref{fig:ablation}(c), initially the model fails to sample any correct trajectories, resulting in a 0\% success rate at the first step. During training, by dynamically incorporating successful trajectories from the pool when all online rollouts fail, the model progressively improves its performance, reaching 46\% by the later steps. This demonstrates that the Experience Pool effectively mitigates the sparse positive signal problem and stabilizes learning on tasks with extremely low natural success rates.

\textbf{High-entropy-driven Step Selection.} 
We evaluate the impact of our high-entropy-driven step selection on agent performance. 
As shown in \Cref{tab:ablation}, compared to the baseline without high-entropy step prioritization, this approach improves success rates from 28.67\% to 68.33\% compared to the baseline, demonstrating its effectiveness.


\textbf{Distribution Alignment.}
We examine the distribution alignment in stabilizing multi-turn VLM RL. As shown in ~\Cref{fig:ablation}(d), incorporating rollout log probabilities as importance weights provides three key benefits: (1) maintains training stability with consistent ~70\% accuracy throughout training, (2) achieves higher peak performance (78\% vs. 55\%), and (3) prevents catastrophic collapse that occurs in the baseline, which drops from 55\% to near 0\% after step 60, a common issue in agent RL. We also observe that our method improves performance over the baseline, reaching 70.55\% in \Cref{tab:ablation}. By reweighting gradients according to the probability ratio between rollout and current distributions, our method effectively mitigates distribution shift, enabling stable learning in long-horizon GUI tasks.



\vspace{-0.2cm}
\section{Conclusion}
\vspace{-0.2cm}
In this paper, we have introduced an efficient reinforcement learning (RL) method to address key challenges in training GUI agents powered by vision-language models (VLMs). 
Our approach overcomes two major limitations in current RL frameworks for GUI tasks.
The proposed decoupled RL framework can speed up the training process, and the data curation scheme can improve the quality of training data for multi-turn agent-environment interactions.
In this case, we significantly improved both GPU and environment utilization and further ensure better agent performance.
Our experimental results on OSWorld demonstrate the efficacy of our approach, achieving a 42.13\% task success rate, outperforming existing open-source models and the baseline by substantial margins. The ablation studies highlight the critical role of the data curation scheme in optimizing the RL process. This work presents a promising direction for improving the efficiency and success of RL-based GUI agents and provides valuable insights for future developments in this field.


\section*{Reproducibility statement}

We have made several efforts to ensure that the results reported in this paper are reproducible.

For the proposed decoupled RL framework, we provide detailed descriptions of the overall architecture and algorithmic components in ~\Cref{sec:dart}, and of the adaptive multi-level data curation and exploration strategies in ~\Cref{sec:data}.

Our experimental setup and evaluation protocols are described in ~\Cref{subsec:setting}, including dataset configurations, environment settings, and performance metrics. The implementation details of the training framework, such as the Trainer, Rollout Service, Environment Cluster, and Data Manager, are provided in ~\Cref{subsec:imple}, with additional specifics in Appendices~\ref{app:bi}–~\ref{app:llm_usage}. 

All ablation studies and efficiency analyses are reported in Sections~\ref{subsec:result}–~\ref{subsec:ablation}, with implementation details and additional examples provided in Appendices~\ref{app:spas}–~\ref{app:fail_case}. These include descriptions of action spaces, system prompts, and extended experimental results, which together provide sufficient information for reproducing the reported performance.

Additionally, source code, configuration files, and pretrained models will be made available to facilitate full reproducibility.

\bibliography{iclr2026_conference}
\bibliographystyle{iclr2026_conference}

\newpage
\appendix
\section{Appendix}
\subsection{Broader Impact}
\label{app:bi}

Our work advances GUI automation through efficient reinforcement learning, offering benefits in accessibility, productivity, and evaluation. DART-GUI can assist users with disabilities, streamline repetitive tasks, and enable more thorough UI validation. By releasing our framework and models openly, we lower the barrier for researchers and developers while reducing environmental impact through efficient training. At the same time, we recognize risks such as unauthorized access or privacy concerns and emphasize the importance of responsible use with proper safeguards.

\subsection{LLM Usage Statement}
\label{app:llm_usage}
We acknowledge the use of LLMs as a writing assistance tool during the preparation of this manuscript. The LLMs were utilized exclusively for improving language quality, including grammar correction, and enhancing clarity. All scientific contributions, including research conceptualization, methodology, experimental design, data analysis, and interpretation of results were conducted solely by the human authors. The LLMs did not generate any original research ideas, hypotheses, or substantive scientific content. The authors assume full responsibility for the accuracy, integrity, and originality of all content presented in this work, including any portions where language was refined with LLM assistance.

\subsection{System Prompt and Action Space}
\label{app:spas}
We follow the system prompt of the baseline model UI-TARS-1.5-7B\citep{qin2025ui}, as shown in ~\Cref{fig:system_prompt}.

\begin{figure}[H]
\centering
\begin{tcolorbox}[
    colback=gray!5!white,
    colframe=gray!75!black,
    title={\textbf{System Prompt for GUI Agent}},
    fonttitle=\bfseries\color{white},
    coltitle=white,
    colbacktitle=gray!75!black,
    width=\textwidth,
    left=2mm,
    right=2mm,
    top=2mm,
    bottom=2mm
]
\small
You are a \textcolor{blue}{\textbf{GUI agent}}. You are given a \textcolor{blue}{\textbf{task}} and your \textcolor{blue}{\textbf{action history}}, with \textcolor{blue}{\textbf{screenshots}}. You need to perform the next action to complete the task.

\vspace{0.5em}
\textbf{\#\# Output Format}
\begin{verbatim}
Thought: ...
Action: ...
\end{verbatim}

\vspace{0.5em}
\textbf{\#\# Action Space}

\begin{ttfamily}\small
\textcolor{blue}{click}(start\_box='<|box\_start|>(x1,y1)<|box\_end|>')\\
\textcolor{blue}{left\_double}(start\_box='<|box\_start|>(x1,y1)<|box\_end|>')\\
\textcolor{blue}{right\_single}(start\_box='<|box\_start|>(x1,y1)<|box\_end|>')\\
\textcolor{blue}{drag}(start\_box='<|box\_start|>(x1,y1)<|box\_end|>',\\
\phantom{drag(}end\_box='<|box\_start|>(x3,y3)<|box\_end|>')\\
\textcolor{blue}{hotkey}(key='')\\
\textcolor{blue}{type}(content='') \# If you want to submit, use "\textbackslash n" at the end\\
\textcolor{blue}{scroll}(start\_box='<|box\_start|>(x1,y1)<|box\_end|>',\\
\phantom{scroll(}direction='down or up or right or left')\\
\textcolor{blue}{wait}() \# Sleep for 5s and take a screenshot\\
\textcolor{blue}{finished}(content='xxx') \# Use escape characters \textbackslash', \textbackslash", \textbackslash n
\end{ttfamily}

\vspace{0.5em}
\textbf{\#\# Note}
\begin{itemize}[leftmargin=*,itemsep=2pt,topsep=0pt]
    \item Use \textcolor{blue}{\texttt{\{language\}}} in \texttt{Thought} part.
    \item Write a \textcolor{blue}{\textbf{small plan}} and finally \textcolor{blue}{\textbf{summarize}} your next action (with its target element) in one sentence in \texttt{Thought} part.
    \item My computer's password is \textcolor{blue}{\texttt{'password'}}, feel free to use it when you need sudo rights.
\end{itemize}

\vspace{0.5em}
\textbf{\#\# User Instruction}\\
\textcolor{blue}{\texttt{\{instruction\}}}
\end{tcolorbox}
\caption{System prompt template for DART-GUI agent with action space definition and output format specifications.}
\label{fig:system_prompt}
\end{figure}

\subsection{More Implementation Details}
\label{app:implementation}
\textbf{Trainer.} 
The training pipeline utilizes Fully Sharded Data Parallel (FSDP)~\citep{zhao2023pytorch} via verl~\citep{sheng2024hybridflow} for distributed training across 8 NVIDIA H100 GPUs. The learning rate is set to $1 \times 10^{-6}$ with a KL divergence regularization coefficient of $\beta = 0.1$. Following DAPO~\citep{yu2025dapo}, dynamic clipping boundaries are configured with $\epsilon_{\text{low}} = 0.2$ and $\epsilon_{\text{high}} = 0.28$. The rollout policy scaling parameter is set to $C = 1$.

\textbf{Rollout Service.}
Model deployment employs vLLM~\citep{kwon2023efficient} as the rollout service, incorporating load balancing mechanisms to distribute workload across devices. \textit{Worker-wise Model Syncing} is implemented to enable non-blocking operation. Each worker is allocated $2$ NVIDIA H100 GPUs. The sampling temperature is set to $1.0$, with a maximum of 30 interaction steps per episode. The initial rollout num is configured as $n_{\text{rollout}} = 8$.

\begin{figure}[htbp]
    \centering
    \includegraphics[width=0.45\textwidth]{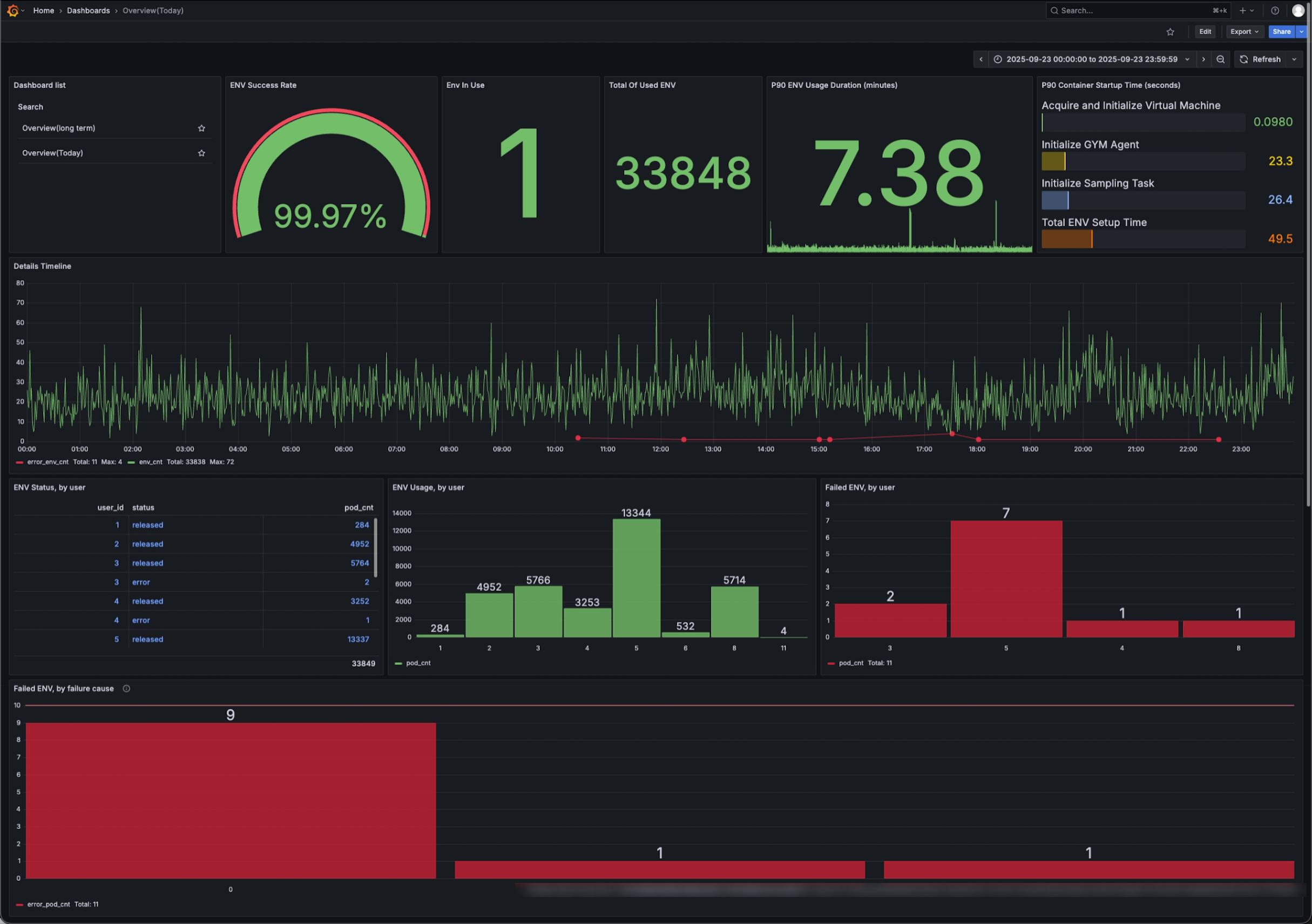}
    \hspace{0.2cm} 
    \includegraphics[width=0.482\textwidth]{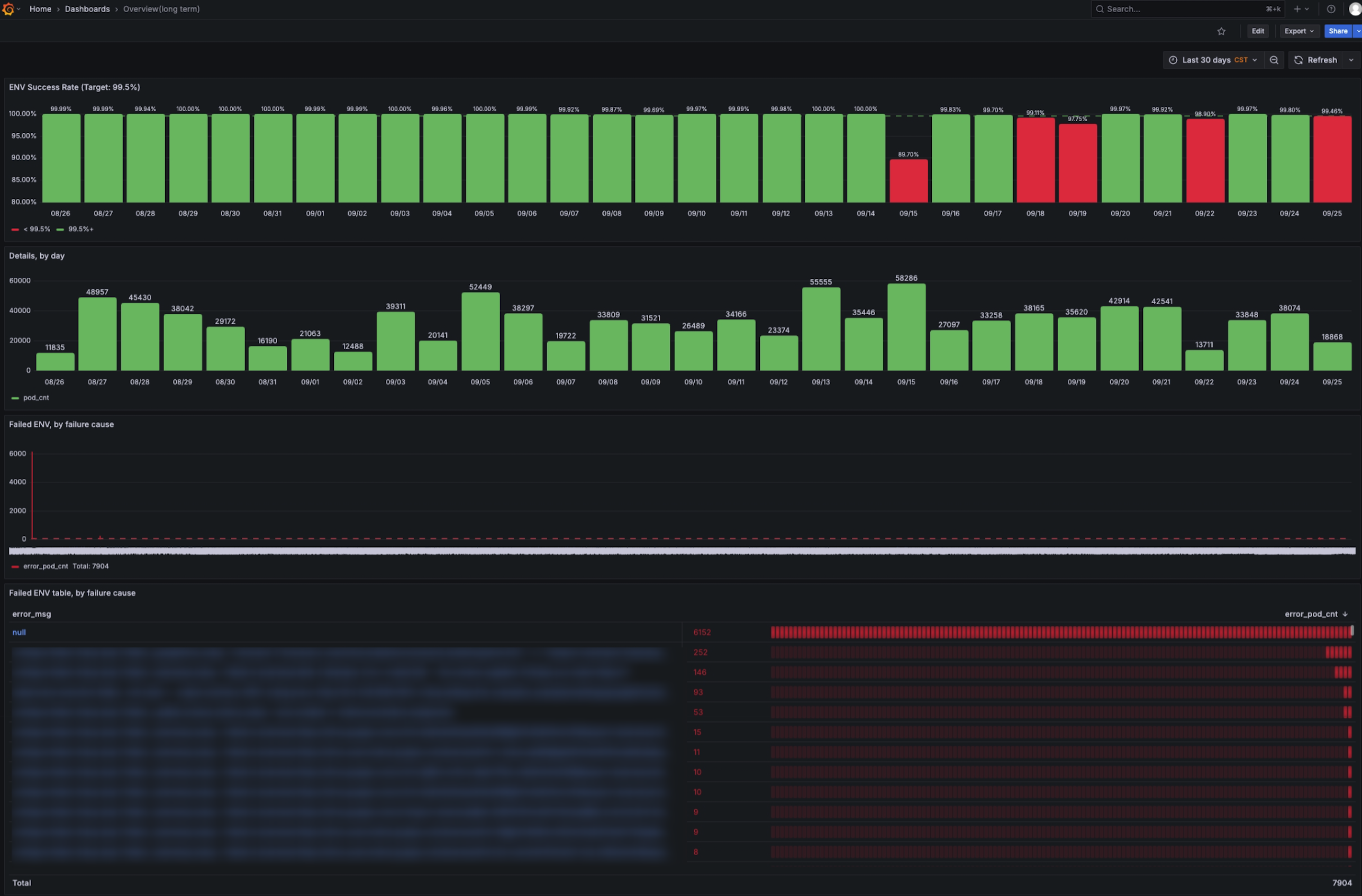}
    \caption{Dashboard of the distributed Ubuntu env cluster.}
    \label{fig:env_dash}
\end{figure}

\textbf{Env Cluster.} We use 
Kubernetes(K8s)~\cite{burns2016borg} to orchestrate $180$ parallel Ubuntu Docker containers serving as environment instances. Each environment operates independently, receiving actions from agents and returning screenshots as observations. ~\Cref{fig:env_dash} shows the dashboard of our cluster.

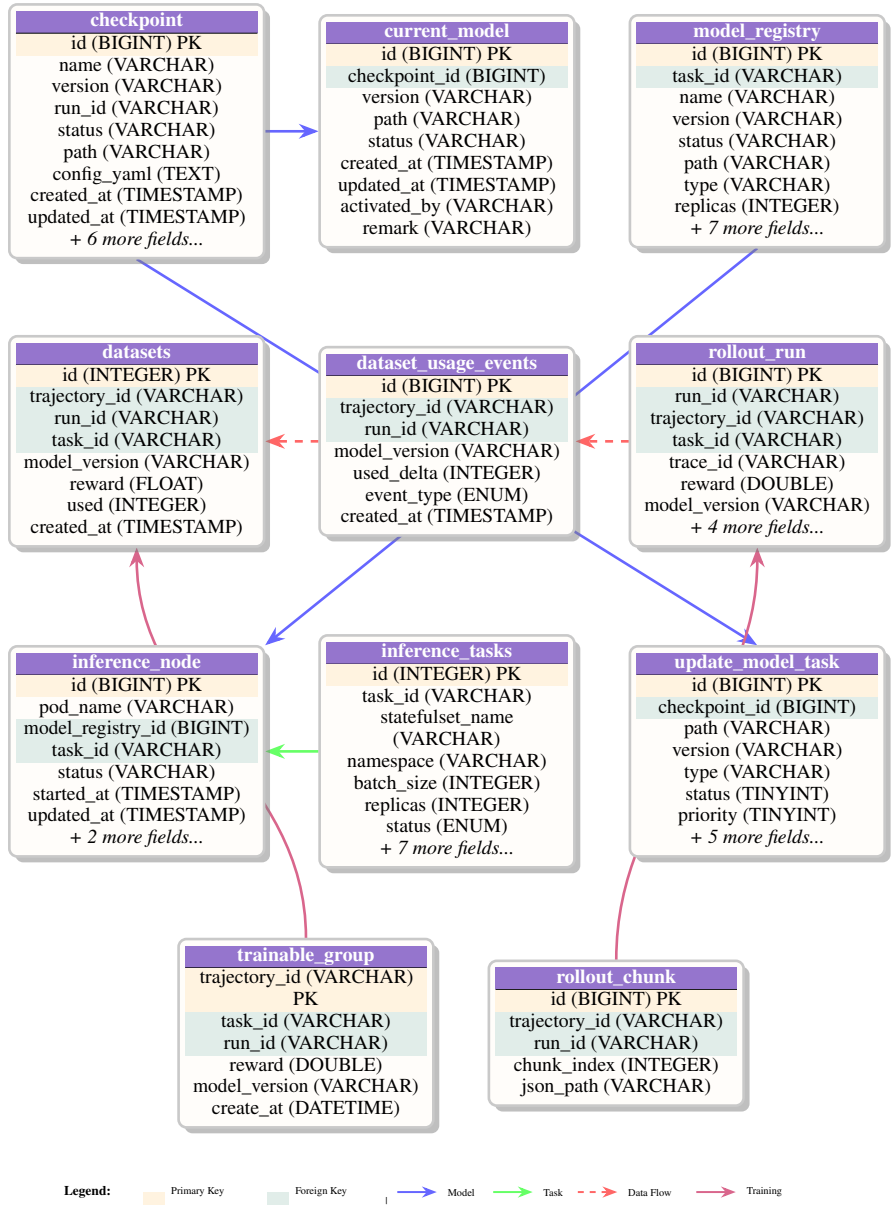
\begin{figure}[htbp]
\centering
\begin{minipage}{\textwidth}
\centering
\begin{tikzpicture}[scale=0.75, every node/.style={scale=0.75}]

\node[table] (checkpoint) at (0,0) {
    \begin{tabular}{>{\centering\arraybackslash}p{4cm}}
    \rowcolor{tableheader}\textcolor{white}{\textbf{checkpoint}} \\
    \hline
    \cellcolor{pkcolor}id (BIGINT) PK \\
    name (VARCHAR) \\
    version (VARCHAR) \\
    run\_id (VARCHAR) \\
    status (VARCHAR) \\
    path (VARCHAR) \\
    config\_yaml (TEXT) \\
    created\_at (TIMESTAMP) \\
    updated\_at (TIMESTAMP) \\
    \textit{+ 6 more fields...}
    \end{tabular}
};

\node[table] (current_model) at (5.5,0) {
    \begin{tabular}{>{\centering\arraybackslash}p{4cm}}
    \rowcolor{tableheader}\textcolor{white}{\textbf{current\_model}} \\
    \hline
    \cellcolor{pkcolor}id (BIGINT) PK \\
    \cellcolor{fkcolor}checkpoint\_id (BIGINT) \\
    version (VARCHAR) \\
    path (VARCHAR) \\
    status (VARCHAR) \\
    created\_at (TIMESTAMP) \\
    updated\_at (TIMESTAMP) \\
    activated\_by (VARCHAR) \\
    remark (VARCHAR)
    \end{tabular}
};

\node[table] (model_registry) at (11,0) {
    \begin{tabular}{>{\centering\arraybackslash}p{4cm}}
    \rowcolor{tableheader}\textcolor{white}{\textbf{model\_registry}} \\
    \hline
    \cellcolor{pkcolor}id (BIGINT) PK \\
    \cellcolor{fkcolor}task\_id (VARCHAR) \\
    name (VARCHAR) \\
    version (VARCHAR) \\
    status (VARCHAR) \\
    path (VARCHAR) \\
    type (VARCHAR) \\
    replicas (INTEGER) \\
    \textit{+ 7 more fields...}
    \end{tabular}
};

\node[table] (datasets) at (0,-5.5) {
    \begin{tabular}{>{\centering\arraybackslash}p{4cm}}
    \rowcolor{tableheader}\textcolor{white}{\textbf{datasets}} \\
    \hline
    \cellcolor{pkcolor}id (INTEGER) PK \\
    \cellcolor{fkcolor}trajectory\_id (VARCHAR) \\
    \cellcolor{fkcolor}run\_id (VARCHAR) \\
    \cellcolor{fkcolor}task\_id (VARCHAR) \\
    model\_version (VARCHAR) \\
    reward (FLOAT) \\
    used (INTEGER) \\
    created\_at (TIMESTAMP)
    \end{tabular}
};

\node[table] (dataset_usage) at (5.5,-5.5) {
    \begin{tabular}{>{\centering\arraybackslash}p{4cm}}
    \rowcolor{tableheader}\textcolor{white}{\textbf{dataset\_usage\_events}} \\
    \hline
    \cellcolor{pkcolor}id (BIGINT) PK \\
    \cellcolor{fkcolor}trajectory\_id (VARCHAR) \\
    \cellcolor{fkcolor}run\_id (VARCHAR) \\
    model\_version (VARCHAR) \\
    used\_delta (INTEGER) \\
    event\_type (ENUM) \\
    created\_at (TIMESTAMP)
    \end{tabular}
};

\node[table] (rollout_run) at (11,-5.5) {
    \begin{tabular}{>{\centering\arraybackslash}p{4cm}}
    \rowcolor{tableheader}\textcolor{white}{\textbf{rollout\_run}} \\
    \hline
    \cellcolor{pkcolor}id (BIGINT) PK \\
    \cellcolor{fkcolor}run\_id (VARCHAR) \\
    \cellcolor{fkcolor}trajectory\_id (VARCHAR) \\
    \cellcolor{fkcolor}task\_id (VARCHAR) \\
    trace\_id (VARCHAR) \\
    reward (DOUBLE) \\
    model\_version (VARCHAR) \\
    \textit{+ 4 more fields...}
    \end{tabular}
};

\node[table] (inference_node) at (0,-11) {
    \begin{tabular}{>{\centering\arraybackslash}p{4cm}}
    \rowcolor{tableheader}\textcolor{white}{\textbf{inference\_node}} \\
    \hline
    \cellcolor{pkcolor}id (BIGINT) PK \\
    pod\_name (VARCHAR) \\
    \cellcolor{fkcolor}model\_registry\_id (BIGINT) \\
    \cellcolor{fkcolor}task\_id (VARCHAR) \\
    status (VARCHAR) \\
    started\_at (TIMESTAMP) \\
    updated\_at (TIMESTAMP) \\
    \textit{+ 2 more fields...}
    \end{tabular}
};

\node[table] (inference_tasks) at (5.5,-11) {
    \begin{tabular}{>{\centering\arraybackslash}p{4cm}}
    \rowcolor{tableheader}\textcolor{white}{\textbf{inference\_tasks}} \\
    \hline
    \cellcolor{pkcolor}id (INTEGER) PK \\
    task\_id (VARCHAR) \\
    statefulset\_name (VARCHAR) \\
    namespace (VARCHAR) \\
    batch\_size (INTEGER) \\
    replicas (INTEGER) \\
    status (ENUM) \\
    \textit{+ 7 more fields...}
    \end{tabular}
};

\node[table] (update_model) at (11,-11) {
    \begin{tabular}{>{\centering\arraybackslash}p{4cm}}
    \rowcolor{tableheader}\textcolor{white}{\textbf{update\_model\_task}} \\
    \hline
    \cellcolor{pkcolor}id (BIGINT) PK \\
    \cellcolor{fkcolor}checkpoint\_id (BIGINT) \\
    path (VARCHAR) \\
    version (VARCHAR) \\
    type (VARCHAR) \\
    status (TINYINT) \\
    priority (TINYINT) \\
    \textit{+ 5 more fields...}
    \end{tabular}
};

\node[table] (trainable) at (3,-16) {
    \begin{tabular}{>{\centering\arraybackslash}p{4cm}}
    \rowcolor{tableheader}\textcolor{white}{\textbf{trainable\_group}} \\
    \hline
    \cellcolor{pkcolor}trajectory\_id (VARCHAR) PK \\
    \cellcolor{fkcolor}task\_id (VARCHAR) \\
    \cellcolor{fkcolor}run\_id (VARCHAR) \\
    reward (DOUBLE) \\
    model\_version (VARCHAR) \\
    create\_at (DATETIME)
    \end{tabular}
};

\node[table] (rollout_chunk) at (8.5,-16) {
    \begin{tabular}{>{\centering\arraybackslash}p{4cm}}
    \rowcolor{tableheader}\textcolor{white}{\textbf{rollout\_chunk}} \\
    \hline
    \cellcolor{pkcolor}id (BIGINT) PK \\
    \cellcolor{fkcolor}trajectory\_id (VARCHAR) \\
    \cellcolor{fkcolor}run\_id (VARCHAR) \\
    chunk\_index (INTEGER) \\
    json\_path (VARCHAR)
    \end{tabular}
};

\begin{scope}[on background layer]
\draw[-{Stealth}, line width=1pt, color=blue!60] 
    (checkpoint.east) -- (current_model.west);

\draw[-{Stealth}, line width=1pt, color=blue!60] 
    (checkpoint.south) -- (update_model.north);

\draw[-{Stealth}, line width=1pt, color=blue!60] 
    (model_registry.south) -- (inference_node.north east);

\draw[-{Stealth}, line width=1pt, color=green!60] 
    (inference_tasks.west) -- (inference_node.east);

\draw[-{Stealth}, line width=1pt, color=red!60, dashed] 
    (rollout_run.west) -- (dataset_usage.east);
\draw[-{Stealth}, line width=1pt, color=red!60, dashed] 
    (dataset_usage.west) -- (datasets.east);

\draw[-{Stealth}, line width=1pt, color=purple!60] 
    (trainable.north) to[out=90,in=270] (datasets.south);
\draw[-{Stealth}, line width=1pt, color=purple!60] 
    (rollout_chunk.north) to[out=90,in=270] (rollout_run.south);
\end{scope}

\node[anchor=north west] at (-1.5,-18.5) {
    \begin{tikzpicture}
    \node[anchor=west, font=\small\bfseries] at (-0.5,0) {Legend:};
    
    \node[rectangle, fill=pkcolor, minimum width=0.5cm, minimum height=0.3cm] at (1,0) {};
    \node[anchor=west, font=\scriptsize] at (1.4,0) {Primary Key};
    
    \node[rectangle, fill=fkcolor, minimum width=0.5cm, minimum height=0.3cm] at (3.2,0) {};
    \node[anchor=west, font=\scriptsize] at (3.6,0) {Foreign Key};

    \node[font=\scriptsize] at (5.2,0) {|};
    
    \draw[-{Stealth}, line width=1pt, color=blue!60] (5.5,0) -- (6.2,0);
    \node[anchor=west, font=\scriptsize] at (6.3,0) {Model};
    
    \draw[-{Stealth}, line width=1pt, color=green!60] (7.2,0) -- (7.9,0);
    \node[anchor=west, font=\scriptsize] at (8,0) {Task};
    
    \draw[-{Stealth}, line width=1pt, color=red!60, dashed] (8.7,0) -- (9.4,0);
    \node[anchor=west, font=\scriptsize] at (9.5,0) {Data Flow};
    
    \draw[-{Stealth}, line width=1pt, color=purple!60] (10.8,0) -- (11.5,0);
    \node[anchor=west, font=\scriptsize] at (11.6,0) {Training};
    \end{tikzpicture}
};

\end{tikzpicture}
\end{minipage}
\caption{Database Schema Visualization showing the relationships between 11 tables. Golden cells indicate primary keys, green cells indicate foreign keys. Blue arrows represent model management relationships, green arrows show task dependencies, red dashed arrows indicate data flow between rollout and dataset tables, and purple arrows represent training data connections.}
\label{fig:database-schema}
\end{figure}

\textbf{Data Manager.}We build a centralized \textit{Data Manager} built on MySQL~\cite{mysql2024} that handles data storage and coordination across the entire training pipeline. The database architecture, illustrated in \Cref{fig:database-schema} and summarized in \Cref{tab:database-categories}, comprises 11 interconnected tables organized into four functional categories. 

For model management, the \textit{Data Manager} tracks model checkpoints and versions through the \texttt{checkpoint}, \texttt{current\_model}, and \texttt{model\_registry} tables, enabling seamless model versioning and deployment. The data management subsystem, consisting of \texttt{datasets}, \texttt{dataset\_usage\_events}, \texttt{rollout\_run}, and \texttt{rollout\_chunk} tables, maintains comprehensive records of trajectories, their usage patterns, and associated rewards. Each trajectory is uniquely identified and linked to its corresponding run, task, and model version.

For the \textit{Trainer}, the \textit{Data Manager} ensures balanced training by monitoring trajectory outcomes through the \texttt{reward} field in multiple tables. When sampling training data, the \textit{Data Manager} guarantees each task contains at least one successful trajectory (positive reward) and one failed trajectory (negative or zero reward). If all sampled trajectories for a given task fail, the \textit{Data Manager} queries the \texttt{datasets} and \texttt{rollout\_run} tables to retrieve positive trajectories from previously collected data, using the \texttt{trajectory\_id} and \texttt{task\_id} as keys to maintain task consistency. The \texttt{dataset\_usage\_events} table tracks these data access patterns, recording each usage with timestamps and model versions to ensure reproducibility. Additionally, the \texttt{trainable\_group} table aggregates trajectories ready for training, while the \texttt{update\_model\_task} table manages the model update pipeline, coordinating between checkpoints and deployment stages.

\begin{table}[H]
\centering
\small
\caption{Database tables categorized by functionality.}
\begin{tabular}{lcp{8cm}}
\hline
\textbf{Category} & \textbf{Table Count} & \textbf{Tables} \\
\hline
Model Management & 3 & checkpoint, current\_model, model\_registry \\
Data Management & 4 & datasets, dataset\_usage\_events, rollout\_run, rollout\_chunk \\
Training & 2 & trainable\_group, update\_model\_task \\
Inference & 2 & inference\_node, inference\_tasks \\
\hline
\textbf{Total} & \textbf{11} & \\
\hline
\end{tabular}

\label{tab:database-categories}
\end{table}

\subsection{Visualization}

\begin{figure}
    \centering
    \includegraphics[width=1.0\linewidth]{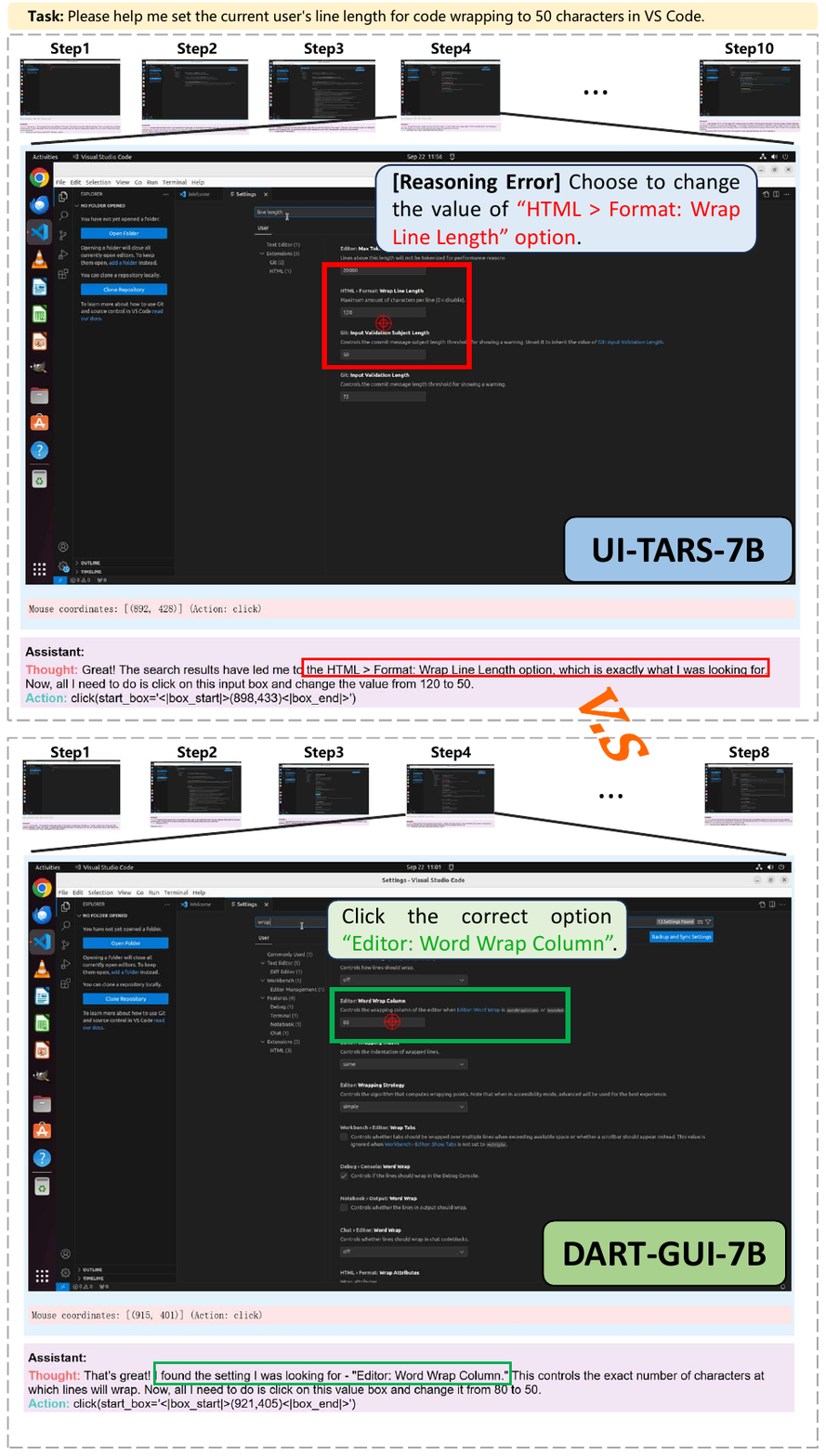}
    \caption{Case study comparing UI-TARS-7B and DART-GUI-7B on configuring line wrapping in VS Code. UI-TARS-7B exhibits a reasoning error by modifying the unrelated HTML > Format: Wrap Line Length option, whereas DART-GUI-7B correctly locates and sets the Editor: Word Wrap Column parameter to the desired value.2}
    \label{fig:case_1}
\end{figure}

\begin{figure}
    \centering
    \includegraphics[width=1.0\linewidth]{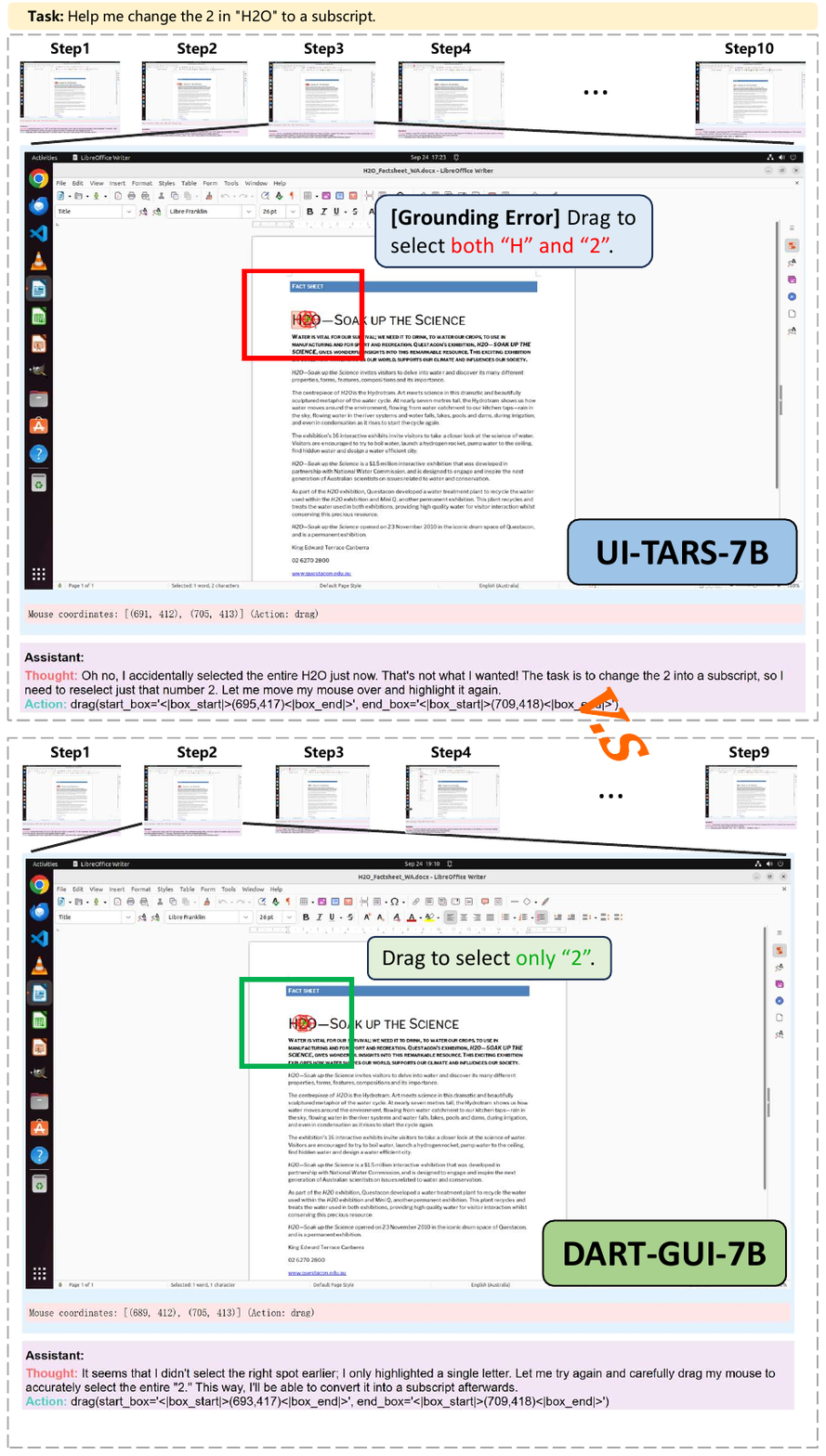}
    \caption{Case study comparing UI-TARS-7B and DART-GUI-7B on editing text in LibreOffice. UI-TARS-7B makes a grounding error by selecting both ``H'' and ``2'' in ``$\mathrm{H_2O}$'', whereas DART-GUI-7B correctly highlights only the ``2'' for conversion into a subscript.}
    \label{fig:case_2}
\end{figure}

\textbf{Key Steps of Tasks.}
We visualize the comparison between the baseline and our model on several tasks, demonstrating that through our RL training, our model can make correct actions at critical steps, thereby achieving successful trajectories. The comparative analysis clearly shows the improvement in decision-making at pivotal moments, as illustrated in \Cref{fig:case_1} and \Cref{fig:case_2}.

\begin{figure}
    \centering
    \includegraphics[width=1.0\linewidth]{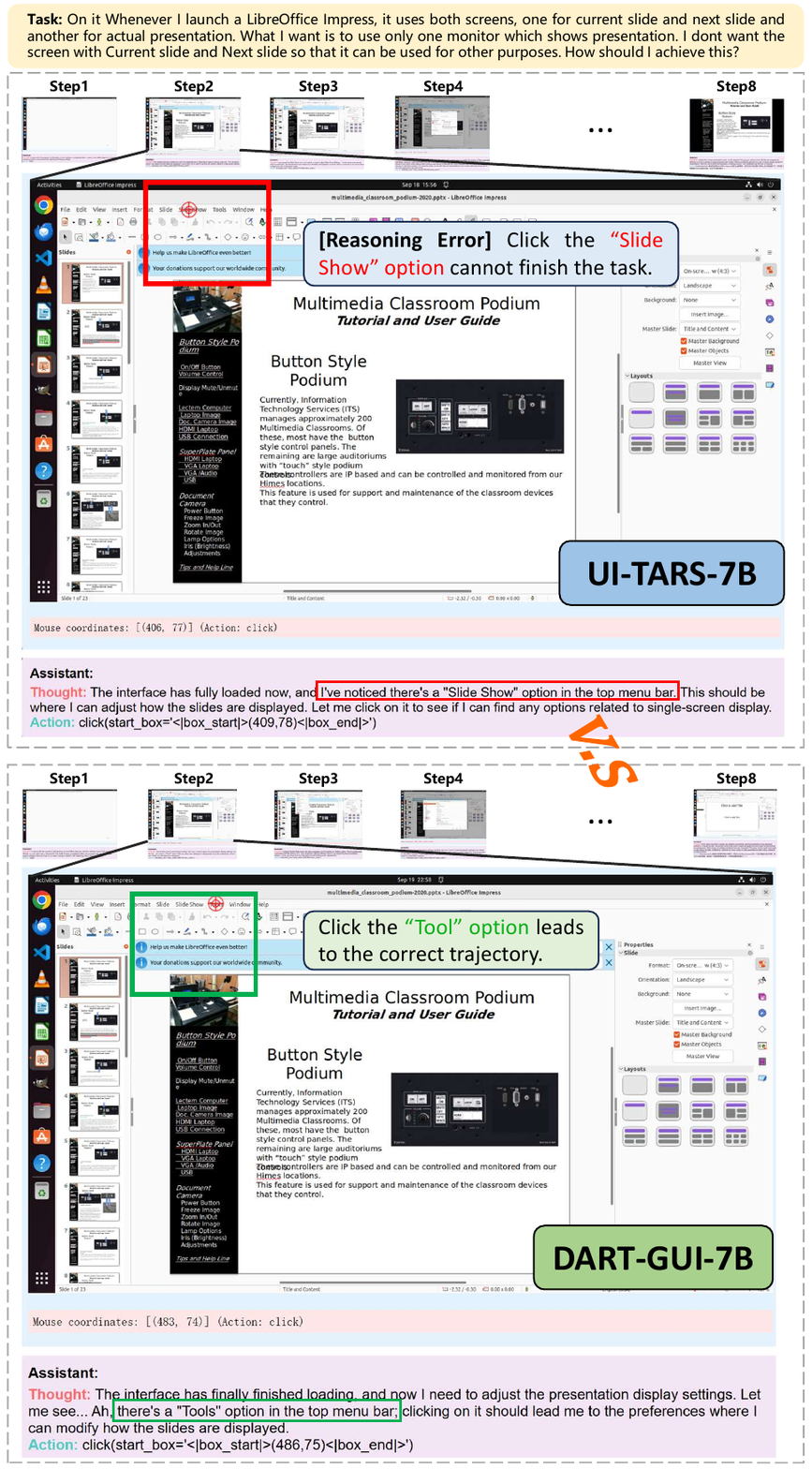}
    \caption{Case study on an extremely difficult LibreOffice Impress task. The task requires configuring dual-slide display settings. The baseline model (top) incorrectly clicks "Slide Show" in the menu, leading to task failure. Our DART-GUI-7B model (bottom), trained with successful trajectories from the trajectory pool, correctly selects "Tools" to access the preferences panel where dual-slide display can be configured. This demonstrates our model's ability to learn from rare successful trajectories and solve previously intractable tasks through RL training.}
    \label{fig:case_plan_1}
\end{figure}

\begin{figure}
    \centering
    \includegraphics[width=1.0\linewidth]{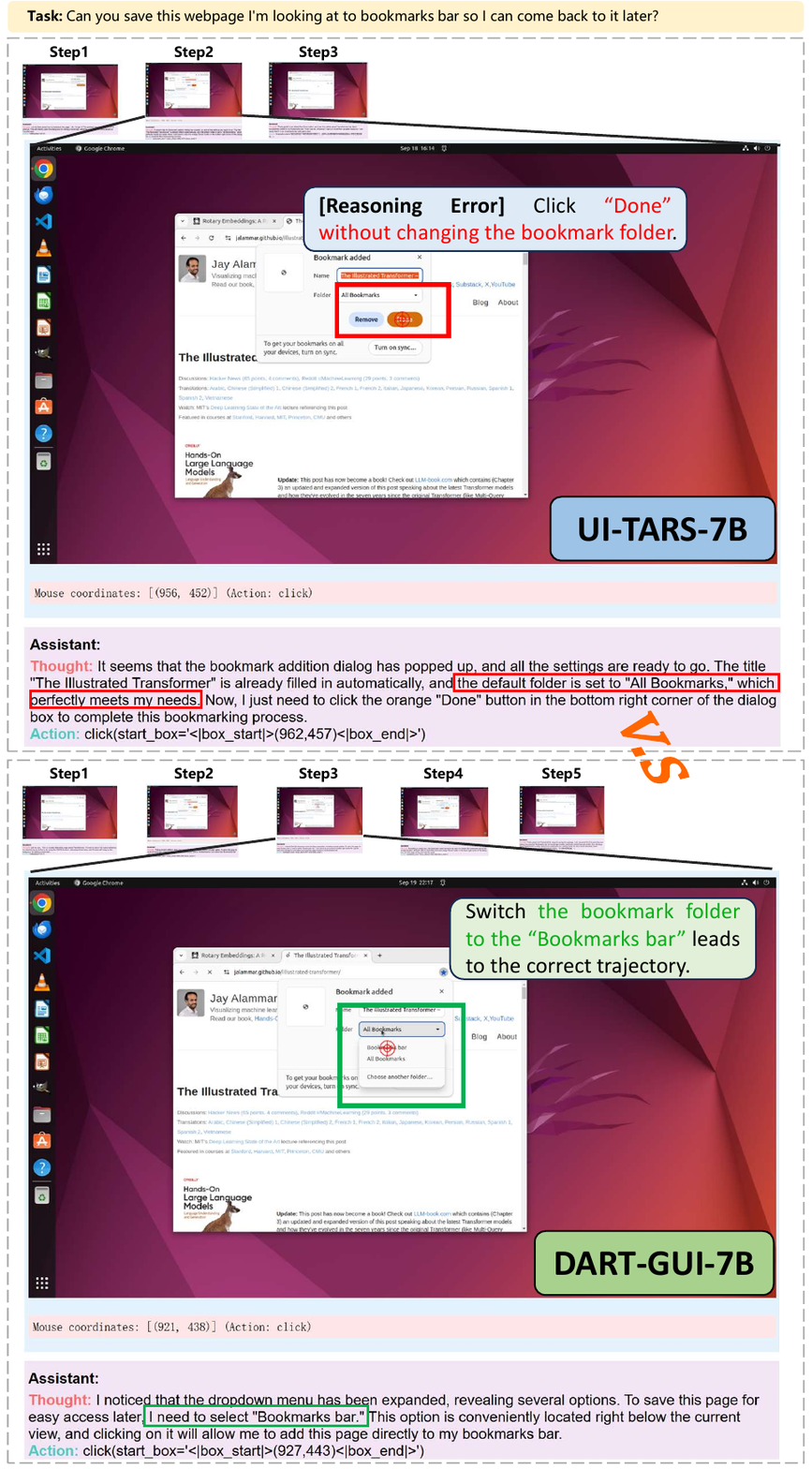}
    \caption{Case study on an extremely difficult bookmark saving task. The task requires saving a webpage to the bookmarks bar for quick access. The baseline model (top) makes a critical error by clicking "Done" without changing the bookmark folder from the default "All Bookmarks" to "Bookmarks bar," resulting in task failure. Our DART-GUI-7B model (bottom) correctly identifies the need to switch the folder dropdown to "Bookmarks bar" before confirming, successfully completing the task. This demonstrates our model's ability to understand subtle but crucial UI requirements that determine task success, learned through RL training on rare successful trajectories.}
    \label{fig:case_plan_2}
\end{figure}

\textbf{Visualization of Extremely Difficult Tasks.}
By leveraging pre-collected successful trajectories from the trajectory pool for extremely difficult tasks (where pass@32 failed), our trained model demonstrates the ability to generate correct trajectories on these challenging tasks. We visualize the critical steps that truly determine the success or failure of trajectories in these tasks and provide detailed analysis of the underlying reasons, as shown in \Cref{fig:case_plan_1} and \Cref{fig:case_plan_2}.

\subsection{Failure Cases}
\label{app:fail_case}
\begin{figure}
    \centering
    \includegraphics[width=1.0\linewidth]{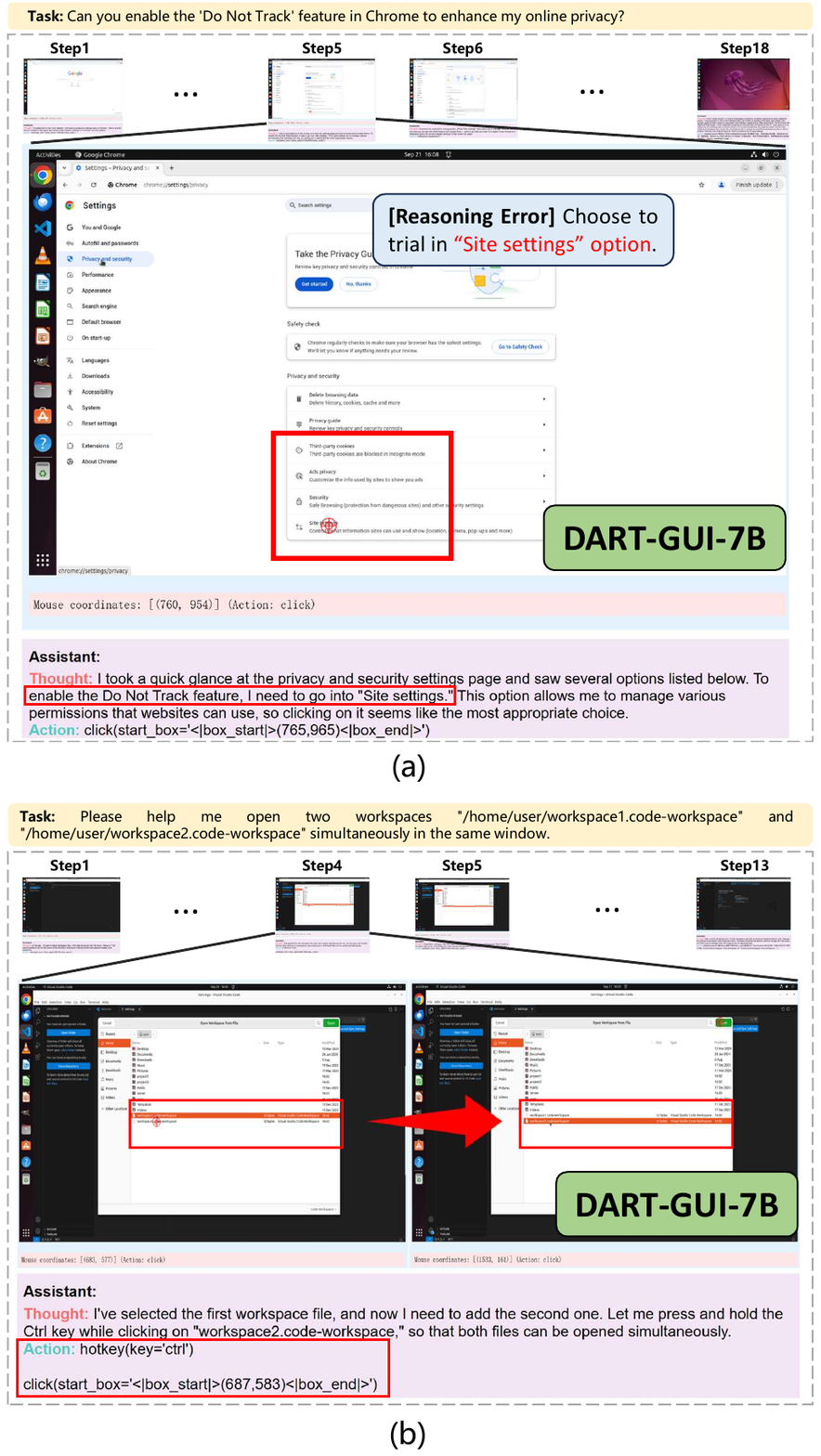}
    \caption{Failure cases of DART-GUI-7B. 
    (a) For the task of enabling the “Do Not Track” feature in Chrome, the model incorrectly clicks the “Site settings” option instead of the “Third-party cookies” option required to access the relevant privacy control. 
    (b) For the task of opening two workspaces simultaneously in VS Code, the model attempts a Ctrl+click sequence, but due to action space limitations, the execution corresponds to sequentially pressing “Ctrl” and clicking without holding “Ctrl”, which deselects the first workspace and leaves only the second one selected.}
    \label{fig:fail_case}
\end{figure}

We also visualize representative failure cases to highlight the limitations of our model. These examples demonstrate situations where DART-GUI-7B makes mistakes at key steps, preventing successful task completion, as illustrated in \Cref{fig:fail_case}.

\end{document}

%% file: math_commands.tex
\usepackage{amsmath,amsfonts,bm}

\def\eqref#1{equation~\ref{#1}}

\def\1{\bm{1}}

\DeclareMathAlphabet{\mathsfit}{\encodingdefault}{\sfdefault}{m}{sl}
\SetMathAlphabet{\mathsfit}{bold}{\encodingdefault}{\sfdefault}{bx}{n}



%% file: config.tex
\usepackage[utf8]{inputenc} 
\usepackage[T1]{fontenc}    
\usepackage{hyperref}       
\usepackage{url}            
\usepackage{booktabs}       
\usepackage{amsfonts}       
\usepackage{nicefrac}       
\usepackage{microtype}      
\usepackage[table]{xcolor}         

\usepackage{tcolorbox}
\usepackage{xcolor}
\usepackage{enumitem}

\usepackage{graphicx}   
\usepackage{amsmath}
\usepackage{amssymb}
\usepackage{acronym}
\usepackage{cleveref}
\usepackage{xspace}
\usepackage{siunitx, tabularx}

\usepackage{algorithm}
\usepackage{algorithmic}
\usepackage{multirow}
\usepackage{subcaption}  
\usepackage{wrapfig}
\usepackage{marvosym}
\usepackage{svg}

\usepackage{graphicx}

\usepackage{tcolorbox}

\definecolor{crimson}{RGB}{220, 20, 60}

\definecolor{customgray}{HTML}{D9D9D9}
\definecolor{customblue}{HTML}{DDE6F7}
\definecolor{customgreen}{HTML}{E2F4E7}

\usepackage{tikz}
\usepackage{array}
\usepackage{xcolor}
\usetikzlibrary{shapes.geometric, arrows.meta, positioning, shadows, backgrounds, calc}

\definecolor{tableheader}{RGB}{149, 117, 205}  
\definecolor{tablebg}{RGB}{255, 253, 250}      
\definecolor{pkcolor}{RGB}{255, 243, 224}      
\definecolor{fkcolor}{RGB}{225, 237, 233}      
\definecolor{bordercolor}{RGB}{201, 201, 201}  

\tikzstyle{table} = [
    rectangle,
    rounded corners=3pt,
    draw=bordercolor,
    line width=1pt,
    minimum width=4.5cm,
    minimum height=1cm,
    text centered,
    drop shadow,
    fill=tablebg
]

\tikzstyle{title} = [
    fill=tableheader,
    text=white,
    font=\bfseries\small,
    minimum height=0.8cm,
    rounded corners=3pt 3pt 0 0
]

\acrodef{dart}[\textsc{DART}]{\underline{D}ecoupled \underline{A}gentic \underline{R}L \underline{T}raining}

%% file: main.bbl
\begin{thebibliography}{57}
\providecommand{\natexlab}[1]{#1}
\providecommand{\url}[1]{\texttt{#1}}
\expandafter\ifx\csname urlstyle\endcsname\relax
  \providecommand{\doi}[1]{doi: #1}\else
  \providecommand{\doi}{doi: \begingroup \urlstyle{rm}\Url}\fi

\bibitem[{Anthropic}(2025{\natexlab{a}})]{anthropic2025a}
{Anthropic}.
\newblock {Claude 3.7 Sonnet and Claude Code}.
\newblock Technical report, Anthropic, 2025{\natexlab{a}}.
\newblock URL \url{https://www.anthropic.com/news/claude-3-7-sonnet}.
\newblock System Card. 12, 13, 14.

\bibitem[{Anthropic}(2025{\natexlab{b}})]{anthropic2025b}
{Anthropic}.
\newblock {Claude-4 Sonnet}.
\newblock Technical report, Anthropic, 2025{\natexlab{b}}.
\newblock URL \url{https://www.anthropic.com/news/claude-4}.
\newblock System Card. 14.

\bibitem[Bai et~al.(2025)Bai, Chen, Liu, Wang, Ge, Song, Dang, Wang, Wang, Tang, et~al.]{bai2025qwen2}
Shuai Bai, Keqin Chen, Xuejing Liu, Jialin Wang, Wenbin Ge, Sibo Song, Kai Dang, Peng Wang, Shijie Wang, Jun Tang, et~al.
\newblock Qwen2. 5-vl technical report.
\newblock \emph{arXiv preprint arXiv:2502.13923}, 2025.

\bibitem[Berner et~al.(2019)Berner, Brockman, Chan, Cheung, D{\k{e}}biak, Dennison, Farhi, Fischer, Hashme, Hesse, et~al.]{berner2019dota}
Christopher Berner, Greg Brockman, Brooke Chan, Vicki Cheung, Przemys{\l}aw D{\k{e}}biak, Christy Dennison, David Farhi, Quirin Fischer, Shariq Hashme, Chris Hesse, et~al.
\newblock Dota 2 with large scale deep reinforcement learning.
\newblock \emph{arXiv preprint arXiv:1912.06680}, 2019.

\bibitem[Burns et~al.(2016)Burns, Grant, Oppenheimer, Brewer, and Wilkes]{burns2016borg}
Brendan Burns, Brian Grant, David Oppenheimer, Eric Brewer, and John Wilkes.
\newblock Borg, omega, and kubernetes.
\newblock In \emph{ACM Queue}, volume~14, pp.\  70--93. ACM, 2016.

\bibitem[Cheng et~al.(2024)Cheng, Sun, Chu, Xu, Li, Zhang, and Wu]{cheng2024seeclick}
Kanzhi Cheng, Qiushi Sun, Yougang Chu, Fangzhi Xu, Yantao Li, Jianbing Zhang, and Zhiyong Wu.
\newblock Seeclick: Harnessing gui grounding for advanced visual gui agents.
\newblock \emph{arXiv preprint arXiv:2401.10935}, 2024.

\bibitem[DeepSeek-AI(2025)]{deepseekai2025deepseekr1}
DeepSeek-AI.
\newblock Deepseek-r1: Incentivizing reasoning capability in llms via reinforcement learning, 2025.
\newblock URL \url{https://arxiv.org/abs/2501.12948}.

\bibitem[Deng et~al.(2023)Deng, Gu, Zheng, Chen, Stevens, Wang, Sun, and Su]{deng2023mind2web}
Xiang Deng, Yu~Gu, Boyuan Zheng, Shijie Chen, Sam Stevens, Boshi Wang, Huan Sun, and Yu~Su.
\newblock Mind2web: Towards a generalist agent for the web.
\newblock \emph{Advances in Neural Information Processing Systems}, 36:\penalty0 28091--28114, 2023.

\bibitem[Fu et~al.(2025{\natexlab{a}})Fu, Su, Zhao, Wang, Wu, Yu, Hu, Shi, Dong, Wang, Chen, Yu, Peng, Li, Huang, Wei, Yu, Xin, Zhao, Gu, Jiang, Zhou, and Wang]{fu2025mano}
Tianyu Fu, Anyang Su, Chenxu Zhao, Hanning Wang, Minghui Wu, Zhe Yu, Fei Hu, Mingjia Shi, Wei Dong, Jiayao Wang, Yuyang Chen, Ruiyang Yu, Siran Peng, Menglin Li, Nan Huang, Haitian Wei, Jiawei Yu, Yi~Xin, Xilin Zhao, Kai Gu, Ping Jiang, Sifan Zhou, and Shuo Wang.
\newblock Mano report.
\newblock \emph{arXiv preprint arXiv:2509.17336}, 2025{\natexlab{a}}.

\bibitem[Fu et~al.(2025{\natexlab{b}})Fu, Gao, Shen, Zhu, Mei, He, Xu, Wei, Mei, Wang, et~al.]{fu2025areal}
Wei Fu, Jiaxuan Gao, Xujie Shen, Chen Zhu, Zhiyu Mei, Chuyi He, Shusheng Xu, Guo Wei, Jun Mei, Jiashu Wang, et~al.
\newblock Areal: A large-scale asynchronous reinforcement learning system for language reasoning.
\newblock \emph{arXiv preprint arXiv:2505.24298}, 2025{\natexlab{b}}.

\bibitem[Gao et~al.(2024{\natexlab{a}})Gao, Du, Zhang, Ma, Han, Zhu, and Li]{gao2024clova}
Zhi Gao, Yuntao Du, Xintong Zhang, Xiaojian Ma, Wenjuan Han, Song-Chun Zhu, and Qing Li.
\newblock Clova: A closed-loop visual assistant with tool usage and update.
\newblock In \emph{Proceedings of the IEEE/CVF conference on computer vision and pattern recognition}, pp.\  13258--13268, 2024{\natexlab{a}}.

\bibitem[Gao et~al.(2024{\natexlab{b}})Gao, Zhang, Li, Ma, Yuan, Fan, Wu, Jia, Zhu, and Li]{gao2024multi}
Zhi Gao, Bofei Zhang, Pengxiang Li, Xiaojian Ma, Tao Yuan, Yue Fan, Yuwei Wu, Yunde Jia, Song-Chun Zhu, and Qing Li.
\newblock Multi-modal agent tuning: Building a vlm-driven agent for efficient tool usage.
\newblock \emph{arXiv preprint arXiv:2412.15606}, 2024{\natexlab{b}}.

\bibitem[Gou et~al.(2024)Gou, Wang, Zheng, Xie, Chang, Shu, Sun, and Su]{gou2024navigating}
Boyu Gou, Ruohan Wang, Boyuan Zheng, Yanan Xie, Cheng Chang, Yiheng Shu, Huan Sun, and Yu~Su.
\newblock Navigating the digital world as humans do: Universal visual grounding for gui agents.
\newblock \emph{arXiv preprint arXiv:2410.05243}, 2024.

\bibitem[Guo et~al.(2025{\natexlab{a}})Guo, Yang, Zhang, Song, Zhang, Xu, Zhu, Ma, Wang, Bi, et~al.]{guo2025deepseek}
Daya Guo, Dejian Yang, Haowei Zhang, Junxiao Song, Ruoyu Zhang, Runxin Xu, Qihao Zhu, Shirong Ma, Peiyi Wang, Xiao Bi, et~al.
\newblock Deepseek-r1: Incentivizing reasoning capability in llms via reinforcement learning.
\newblock \emph{arXiv preprint arXiv:2501.12948}, 2025{\natexlab{a}}.

\bibitem[Guo et~al.(2025{\natexlab{b}})Guo, Wu, Zhu, Leng, Shi, Chen, Fan, Wang, Jiang, Wang, et~al.]{guo2025seed1}
Dong Guo, Faming Wu, Feida Zhu, Fuxing Leng, Guang Shi, Haobin Chen, Haoqi Fan, Jian Wang, Jianyu Jiang, Jiawei Wang, et~al.
\newblock Seed1. 5-vl technical report.
\newblock \emph{arXiv preprint arXiv:2505.07062}, 2025{\natexlab{b}}.

\bibitem[Gur et~al.(2023)Gur, Furuta, Huang, Safdari, Matsuo, Eck, and Faust]{gur2023real}
Izzeddin Gur, Hiroki Furuta, Austin Huang, Mustafa Safdari, Yutaka Matsuo, Douglas Eck, and Aleksandra Faust.
\newblock A real-world webagent with planning, long context understanding, and program synthesis.
\newblock \emph{arXiv preprint arXiv:2307.12856}, 2023.

\bibitem[He et~al.(2024)He, Yao, Ma, Yu, Dai, Zhang, Lan, and Yu]{he2024webvoyager}
Hongliang He, Wenlin Yao, Kaixin Ma, Wenhao Yu, Yong Dai, Hongming Zhang, Zhenzhong Lan, and Dong Yu.
\newblock Webvoyager: Building an end-to-end web agent with large multimodal models.
\newblock \emph{arXiv preprint arXiv:2401.13919}, 2024.

\bibitem[Hong et~al.(2024)Hong, Wang, Lv, Xu, Yu, Ji, Wang, Wang, Dong, Ding, et~al.]{hong2024cogagent}
Wenyi Hong, Weihan Wang, Qingsong Lv, Jiazheng Xu, Wenmeng Yu, Junhui Ji, Yan Wang, Zihan Wang, Yuxiao Dong, Ming Ding, et~al.
\newblock Cogagent: A visual language model for gui agents.
\newblock In \emph{Proceedings of the IEEE/CVF Conference on Computer Vision and Pattern Recognition}, pp.\  14281--14290, 2024.

\bibitem[Kwon et~al.(2023)Kwon, Li, Zhuang, Sheng, Zheng, Yu, Gonzalez, Zhang, and Stoica]{kwon2023efficient}
Woosuk Kwon, Zhuohan Li, Siyuan Zhuang, Ying Sheng, Lianmin Zheng, Cody~Hao Yu, Joseph~E. Gonzalez, Hao Zhang, and Ion Stoica.
\newblock Efficient memory management for large language model serving with pagedattention.
\newblock In \emph{Proceedings of the ACM SIGOPS 29th Symposium on Operating Systems Principles}, 2023.

\bibitem[Lai et~al.(2024)Lai, Liu, Iong, Yao, Chen, Shen, Yu, Zhang, Zhang, Dong, et~al.]{lai2024autowebglm}
Hanyu Lai, Xiao Liu, Iat~Long Iong, Shuntian Yao, Yuxuan Chen, Pengbo Shen, Hao Yu, Hanchen Zhang, Xiaohan Zhang, Yuxiao Dong, et~al.
\newblock Autowebglm: A large language model-based web navigating agent.
\newblock In \emph{Proceedings of the 30th ACM SIGKDD Conference on Knowledge Discovery and Data Mining}, pp.\  5295--5306, 2024.

\bibitem[Lai et~al.(2025)Lai, Liu, Zhao, Xu, Zhang, Jing, Ren, Yao, Dong, and Tang]{lai2025computerrl}
Hanyu Lai, Xiao Liu, Yanxiao Zhao, Han Xu, Hanchen Zhang, Bohao Jing, Yanyu Ren, Shuntian Yao, Yuxiao Dong, and Jie Tang.
\newblock Computerrl: Scaling end-to-end online reinforcement learning for computer use agents.
\newblock \emph{arXiv preprint arXiv:2508.14040}, 2025.

\bibitem[Li et~al.(2025)Li, Gao, Zhang, Mi, Ma, Shi, Yuan, Wu, Jia, Zhu, et~al.]{li2025iterative}
Pengxiang Li, Zhi Gao, Bofei Zhang, Yapeng Mi, Xiaojian Ma, Chenrui Shi, Tao Yuan, Yuwei Wu, Yunde Jia, Song-Chun Zhu, et~al.
\newblock Iterative tool usage exploration for multimodal agents via step-wise preference tuning.
\newblock \emph{arXiv preprint arXiv:2504.21561}, 2025.

\bibitem[Lin et~al.(2024)Lin, Li, Gao, Yang, Bai, Lei, Wang, and Shou]{lin2024showui}
Kevin~Qinghong Lin, Linjie Li, Difei Gao, Zhengyuan Yang, Zechen Bai, Weixian Lei, Lijuan Wang, and Mike~Zheng Shou.
\newblock Showui: One vision-language-action model for generalist gui agent.
\newblock In \emph{NeurIPS 2024 Workshop on Open-World Agents}, volume~1, 2024.

\bibitem[Liu et~al.(2024)Liu, Qin, Liang, Dong, Lai, Zhang, Zhao, Iong, Sun, Wang, et~al.]{liu2024autoglm}
Xiao Liu, Bo~Qin, Dongzhu Liang, Guang Dong, Hanyu Lai, Hanchen Zhang, Hanlin Zhao, Iat~Long Iong, Jiadai Sun, Jiaqi Wang, et~al.
\newblock Autoglm: Autonomous foundation agents for guis.
\newblock \emph{arXiv preprint arXiv:2411.00820}, 2024.

\bibitem[Liu et~al.(2025)Liu, Li, Xie, Hu, Han, Zhang, Yang, and Wu]{liu2025infigui}
Yuhang Liu, Pengxiang Li, Congkai Xie, Xavier Hu, Xiaotian Han, Shengyu Zhang, Hongxia Yang, and Fei Wu.
\newblock Infigui-r1: Advancing multimodal gui agents from reactive actors to deliberative reasoners.
\newblock \emph{arXiv preprint arXiv:2504.14239}, 2025.

\bibitem[Lu et~al.(2025)Lu, Zhong, Liu, Fu, and Jia]{lu2025arpo}
Fanbin Lu, Zhisheng Zhong, Shu Liu, Chi-Wing Fu, and Jiaya Jia.
\newblock Arpo: End-to-end policy optimization for gui agents with experience replay.
\newblock \emph{arXiv preprint arXiv:2505.16282}, 2025.

\bibitem[Lu et~al.(2024)Lu, Yang, Shen, and Awadallah]{lu2024omniparser}
Yadong Lu, Jianwei Yang, Yelong Shen, and Ahmed Awadallah.
\newblock Omniparser for pure vision based gui agent.
\newblock \emph{arXiv preprint arXiv:2408.00203}, 2024.

\bibitem[Luo et~al.(2025)Luo, Wang, He, and Xia]{luo2025gui}
Run Luo, Lu~Wang, Wanwei He, and Xiaobo Xia.
\newblock Gui-r1: A generalist r1-style vision-language action model for gui agents.
\newblock \emph{arXiv preprint arXiv:2504.10458}, 2025.

\bibitem[OpenAI(2025)]{OpenAI_CUA_2025}
OpenAI.
\newblock Computer-using agent (cua): Model for gui interaction and task automation.
\newblock Research preview / API documentation, March 2025.
\newblock URL \url{https://openai.com/index/computer-using-agent/}.
\newblock Powering Operator; “computer-use-preview” model; accessed via Responses API; performance: OSWorld 38.1\% for computer tasks, WebArena 58.1\%, WebVoyager 87\% :contentReference[oaicite:0]{index=0}.

\bibitem[{OpenAI}(2025)]{openai2025b}
{OpenAI}.
\newblock {OpenAI O3 and O4-Mini System Card}.
\newblock Technical report, OpenAI, 2025.
\newblock URL \url{https://cdn.openai.com/pdf/2221c875-02dc-4789-800b-e7758f3722c1/o3-and-o4-mini-system-card.pdf}.
\newblock System Card. 14.

\bibitem[{Oracle Corporation}(2024)]{mysql2024}
{Oracle Corporation}.
\newblock {MySQL Database Management System}, 2024.
\newblock URL \url{https://www.mysql.com/}.
\newblock Accessed: 2024-09-24.

\bibitem[Ouyang et~al.(2022)Ouyang, Wu, Jiang, Almeida, Wainwright, Mishkin, Zhang, Agarwal, Slama, Ray, et~al.]{ouyang2022training}
Long Ouyang, Jeffrey Wu, Xu~Jiang, Diogo Almeida, Carroll Wainwright, Pamela Mishkin, Chong Zhang, Sandhini Agarwal, Katarina Slama, Alex Ray, et~al.
\newblock Training language models to follow instructions with human feedback.
\newblock \emph{Advances in neural information processing systems}, 35:\penalty0 27730--27744, 2022.

\bibitem[Qin et~al.(2025)Qin, Ye, Fang, Wang, Liang, Tian, Zhang, Li, Li, Huang, et~al.]{qin2025ui}
Yujia Qin, Yining Ye, Junjie Fang, Haoming Wang, Shihao Liang, Shizuo Tian, Junda Zhang, Jiahao Li, Yunxin Li, Shijue Huang, et~al.
\newblock Ui-tars: Pioneering automated gui interaction with native agents.
\newblock \emph{arXiv preprint arXiv:2501.12326}, 2025.

\bibitem[Rafailov et~al.(2023)Rafailov, Sharma, Mitchell, Manning, Ermon, and Finn]{rafailov2023direct}
Rafael Rafailov, Archit Sharma, Eric Mitchell, Christopher~D Manning, Stefano Ermon, and Chelsea Finn.
\newblock Direct preference optimization: Your language model is secretly a reward model.
\newblock \emph{Advances in neural information processing systems}, 36:\penalty0 53728--53741, 2023.

\bibitem[Shao et~al.(2024)Shao, Wang, Zhu, Xu, Song, Bi, Zhang, Zhang, Li, et~al.]{shao2024deepseekmath}
Zhihong Shao, Peiyi Wang, Qihao Zhu, Runxin Xu, Junxiao Song, Xiao Bi, Haowei Zhang, Mingchuan Zhang, YK~Li, et~al.
\newblock Deepseekmath: Pushing the limits of mathematical reasoning in open language models.
\newblock \emph{arXiv preprint arXiv:2402.03300}, 2024.

\bibitem[Shen et~al.(2025)Shen, Liu, Li, Fang, Ma, Liao, Shen, Zhang, Zhao, Zhang, et~al.]{shen2025vlm}
Haozhan Shen, Peng Liu, Jingcheng Li, Chunxin Fang, Yibo Ma, Jiajia Liao, Qiaoli Shen, Zilun Zhang, Kangjia Zhao, Qianqian Zhang, et~al.
\newblock Vlm-r1: A stable and generalizable r1-style large vision-language model.
\newblock \emph{arXiv preprint arXiv:2504.07615}, 2025.

\bibitem[Sheng et~al.(2024)Sheng, Zhang, Ye, Wu, Zhang, Zhang, Peng, Lin, and Wu]{sheng2024hybridflow}
Guangming Sheng, Chi Zhang, Zilingfeng Ye, Xibin Wu, Wang Zhang, Ru~Zhang, Yanghua Peng, Haibin Lin, and Chuan Wu.
\newblock Hybridflow: A flexible and efficient rlhf framework.
\newblock \emph{arXiv preprint arXiv: 2409.19256}, 2024.

\bibitem[Tang et~al.(2025)Tang, Li, Cheng, Huo, Wang, Yan, Huang, Jing, and Duan]{tang2025sea}
Liang Tang, Shuxian Li, Yuhao Cheng, Yukang Huo, Zhepeng Wang, Yiqiang Yan, Kaer Huang, Yanzhe Jing, and Tiaonan Duan.
\newblock Sea: Self-evolution agent with step-wise reward for computer use.
\newblock \emph{arXiv preprint arXiv:2508.04037}, 2025.

\bibitem[Vinyals et~al.(2019)Vinyals, Babuschkin, Czarnecki, Mathieu, Dudzik, Chung, Choi, Powell, Ewalds, Georgiev, et~al.]{vinyals2019grandmaster}
Oriol Vinyals, Igor Babuschkin, Wojciech~M Czarnecki, Micha{\"e}l Mathieu, Andrew Dudzik, Junyoung Chung, David~H Choi, Richard Powell, Timo Ewalds, Petko Georgiev, et~al.
\newblock Grandmaster level in starcraft ii using multi-agent reinforcement learning.
\newblock \emph{nature}, 575\penalty0 (7782):\penalty0 350--354, 2019.

\bibitem[Wang et~al.(2025{\natexlab{a}})Wang, Zou, Song, Feng, Fang, Lu, Liu, Luo, Liang, Huang, et~al.]{wang2025ui}
Haoming Wang, Haoyang Zou, Huatong Song, Jiazhan Feng, Junjie Fang, Junting Lu, Longxiang Liu, Qinyu Luo, Shihao Liang, Shijue Huang, et~al.
\newblock Ui-tars-2 technical report: Advancing gui agent with multi-turn reinforcement learning.
\newblock \emph{arXiv preprint arXiv:2509.02544}, 2025{\natexlab{a}}.

\bibitem[Wang et~al.(2025{\natexlab{b}})Wang, Yu, Gao, Zheng, Liu, Lu, Dang, Chen, Yang, Zhang, et~al.]{wang2025beyond}
Shenzhi Wang, Le~Yu, Chang Gao, Chujie Zheng, Shixuan Liu, Rui Lu, Kai Dang, Xionghui Chen, Jianxin Yang, Zhenru Zhang, et~al.
\newblock Beyond the 80/20 rule: High-entropy minority tokens drive effective reinforcement learning for llm reasoning.
\newblock \emph{arXiv preprint arXiv:2506.01939}, 2025{\natexlab{b}}.

\bibitem[Wang et~al.()Wang, Wu, Liu, HAO, Wang, and Shao]{wangdistrl}
Taiyi Wang, Zhihao Wu, Jianheng Liu, Jianye HAO, Jun Wang, and Kun Shao.
\newblock Distrl: An asynchronous distributed reinforcement learning framework for on-device control agent.
\newblock In \emph{The Thirteenth International Conference on Learning Representations}.

\bibitem[Wang et~al.(2025{\natexlab{c}})Wang, Xiong, Chen, Gao, Guo, He, Huang, Liu, Li, Li, et~al.]{wang2025reinforcement}
Weixun Wang, Shaopan Xiong, Gengru Chen, Wei Gao, Sheng Guo, Yancheng He, Ju~Huang, Jiaheng Liu, Zhendong Li, Xiaoyang Li, et~al.
\newblock Reinforcement learning optimization for large-scale learning: An efficient and user-friendly scaling library.
\newblock \emph{arXiv preprint arXiv:2506.06122}, 2025{\natexlab{c}}.

\bibitem[Wang et~al.(2025{\natexlab{d}})Wang, Wang, Lu, Yang, Xie, Wang, Deng, Guo, Xu, Wu, Shen, Li, Li, Li, Chen, Zheng, Li, Lei, Cao, Fu, Shin, Shin, Hu, Wang, Chen, Ye, Zhang, Du, Hu, Chen, Zhou, Yao, Chen, Gu, Wang, Wang, Yang, Zhong, Sung, Charles, Yang, and Yu]{wang2025opencuaopenfoundationscomputeruse}
Xinyuan Wang, Bowen Wang, Dunjie Lu, Junlin Yang, Tianbao Xie, Junli Wang, Jiaqi Deng, Xiaole Guo, Yiheng Xu, Chen~Henry Wu, Zhennan Shen, Zhuokai Li, Ryan Li, Xiaochuan Li, Junda Chen, Boyuan Zheng, Peihang Li, Fangyu Lei, Ruisheng Cao, Yeqiao Fu, Dongchan Shin, Martin Shin, Jiarui Hu, Yuyan Wang, Jixuan Chen, Yuxiao Ye, Danyang Zhang, Dikang Du, Hao Hu, Huarong Chen, Zaida Zhou, Haotian Yao, Ziwei Chen, Qizheng Gu, Yipu Wang, Heng Wang, Diyi Yang, Victor Zhong, Flood Sung, Y.~Charles, Zhilin Yang, and Tao Yu.
\newblock Opencua: Open foundations for computer-use agents, 2025{\natexlab{d}}.
\newblock URL \url{https://arxiv.org/abs/2508.09123}.

\bibitem[Wang et~al.(2025{\natexlab{e}})Wang, Wang, Lu, Yang, Xie, Wang, Deng, Guo, Xu, Wu, et~al.]{wang2025opencua}
Xinyuan Wang, Bowen Wang, Dunjie Lu, Junlin Yang, Tianbao Xie, Junli Wang, Jiaqi Deng, Xiaole Guo, Yiheng Xu, Chen~Henry Wu, et~al.
\newblock Opencua: Open foundations for computer-use agents.
\newblock \emph{arXiv preprint arXiv:2508.09123}, 2025{\natexlab{e}}.

\bibitem[Wang et~al.(2024)Wang, Sun, Zhang, Xian, Biyik, Held, and Erickson]{wang2024rl}
Yufei Wang, Zhanyi Sun, Jesse Zhang, Zhou Xian, Erdem Biyik, David Held, and Zackory Erickson.
\newblock Rl-vlm-f: reinforcement learning from vision language foundation model feedback.
\newblock In \emph{Proceedings of the 41st International Conference on Machine Learning}, pp.\  51484--51501, 2024.

\bibitem[Wu et~al.(2024)Wu, Wu, Xu, Wang, Sun, Jia, Cheng, Ding, Chen, Liang, et~al.]{wu2024atlas}
Zhiyong Wu, Zhenyu Wu, Fangzhi Xu, Yian Wang, Qiushi Sun, Chengyou Jia, Kanzhi Cheng, Zichen Ding, Liheng Chen, Paul~Pu Liang, et~al.
\newblock Os-atlas: A foundation action model for generalist gui agents.
\newblock \emph{arXiv preprint arXiv:2410.23218}, 2024.

\bibitem[Xi et~al.(2025)Xi, Huang, Liao, Huang, Guo, Liu, Zheng, Ye, Zhang, Chen, et~al.]{xi2025agentgym}
Zhiheng Xi, Jixuan Huang, Chenyang Liao, Baodai Huang, Honglin Guo, Jiaqi Liu, Rui Zheng, Junjie Ye, Jiazheng Zhang, Wenxiang Chen, et~al.
\newblock Agentgym-rl: Training llm agents for long-horizon decision making through multi-turn reinforcement learning.
\newblock \emph{arXiv preprint arXiv:2509.08755}, 2025.

\bibitem[Xie et~al.(2024)Xie, Zhang, Chen, Li, Zhao, Cao, Hua, Cheng, Shin, Lei, et~al.]{xie2024osworld}
Tianbao Xie, Danyang Zhang, Jixuan Chen, Xiaochuan Li, Siheng Zhao, Ruisheng Cao, Toh~J Hua, Zhoujun Cheng, Dongchan Shin, Fangyu Lei, et~al.
\newblock Osworld: Benchmarking multimodal agents for open-ended tasks in real computer environments.
\newblock \emph{Advances in Neural Information Processing Systems}, 37:\penalty0 52040--52094, 2024.

\bibitem[Xu et~al.(2024{\natexlab{a}})Xu, Wang, Wang, Lu, Xie, Saha, Sahoo, Yu, and Xiong]{xu2024aguvis}
Yiheng Xu, Zekun Wang, Junli Wang, Dunjie Lu, Tianbao Xie, Amrita Saha, Doyen Sahoo, Tao Yu, and Caiming Xiong.
\newblock Aguvis: Unified pure vision agents for autonomous gui interaction.
\newblock \emph{arXiv preprint arXiv:2412.04454}, 2024{\natexlab{a}}.

\bibitem[Xu et~al.(2024{\natexlab{b}})Xu, Dai, Wei, Yin, and Hu]{xu2024gspo}
Zheng Xu, Xu~Dai, Shaojun Wei, Shouyi Yin, and Yang Hu.
\newblock Gspo: A graph substitution and parallelization joint optimization framework for dnn inference.
\newblock In \emph{Proceedings of the 61st ACM/IEEE Design Automation Conference}, pp.\  1--6, 2024{\natexlab{b}}.

\bibitem[Yang et~al.(2025)Yang, Su, Liu, Dong, Yu, Su, Wang, Liu, Zhu, Li, et~al.]{yang2025zerogui}
Chenyu Yang, Shiqian Su, Shi Liu, Xuan Dong, Yue Yu, Weijie Su, Xuehui Wang, Zhaoyang Liu, Jinguo Zhu, Hao Li, et~al.
\newblock Zerogui: Automating online gui learning at zero human cost.
\newblock \emph{arXiv preprint arXiv:2505.23762}, 2025.

\bibitem[Yao et~al.(2025)Yao, Liu, Zhang, Dong, Shang, and Gao]{yao2025offpolicy}
Feng Yao, Liyuan Liu, Dinghuai Zhang, Chengyu Dong, Jingbo Shang, and Jianfeng Gao.
\newblock Your efficient rl framework secretly brings you off-policy rl training, August 2025.
\newblock URL \url{https://fengyao.notion.site/off-policy-rl}.

\bibitem[Ye et~al.(2025)Ye, Zhang, Xu, Liu, Wang, Zhu, Zheng, Gao, Cao, Lu, et~al.]{ye2025mobile}
Jiabo Ye, Xi~Zhang, Haiyang Xu, Haowei Liu, Junyang Wang, Zhaoqing Zhu, Ziwei Zheng, Feiyu Gao, Junjie Cao, Zhengxi Lu, et~al.
\newblock Mobile-agent-v3: Foundamental agents for gui automation.
\newblock \emph{arXiv preprint arXiv:2508.15144}, 2025.

\bibitem[Yu et~al.(2025)Yu, Zhang, Zhu, Yuan, Zuo, Yue, Dai, Fan, Liu, Liu, et~al.]{yu2025dapo}
Qiying Yu, Zheng Zhang, Ruofei Zhu, Yufeng Yuan, Xiaochen Zuo, Yu~Yue, Weinan Dai, Tiantian Fan, Gaohong Liu, Lingjun Liu, et~al.
\newblock Dapo: An open-source llm reinforcement learning system at scale.
\newblock \emph{arXiv preprint arXiv:2503.14476}, 2025.

\bibitem[Zhang et~al.(2025)Zhang, Shang, Gao, Zhang, Xie, Ma, Yuan, Wu, Zhu, and Li]{zhang2025tongui}
Bofei Zhang, Zirui Shang, Zhi Gao, Wang Zhang, Rui Xie, Xiaojian Ma, Tao Yuan, Xinxiao Wu, Song-Chun Zhu, and Qing Li.
\newblock Tongui: Building generalized gui agents by learning from multimodal web tutorials.
\newblock \emph{arXiv preprint arXiv:2504.12679}, 2025.

\bibitem[Zhao et~al.(2023)Zhao, Gu, Varma, Luo, Huang, Xu, Wright, Shojanazeri, Ott, Shleifer, et~al.]{zhao2023pytorch}
Yanli Zhao, Andrew Gu, Rohan Varma, Liang Luo, Chien-Chin Huang, Min Xu, Less Wright, Hamid Shojanazeri, Myle Ott, Sam Shleifer, et~al.
\newblock Pytorch fsdp: experiences on scaling fully sharded data parallel.
\newblock \emph{arXiv preprint arXiv:2304.11277}, 2023.

\end{thebibliography}
